\documentclass[sigconf]{acmart}
\usepackage{multirow}
\usepackage{makecell}

\AtBeginDocument{%
  }

\copyrightyear{2023}
\acmYear{2023}
\setcopyright{acmlicensed}\acmConference[CIKM '23]{Proceedings of the 32nd ACM International Conference on Information and Knowledge Management}{October 21--25, 2023}{Birmingham, United Kingdom}
\acmBooktitle{Proceedings of the 32nd ACM International Conference on Information and Knowledge Management (CIKM '23), October 21--25, 2023, Birmingham, United Kingdom}
\acmPrice{15.00}
\acmDOI{10.1145/3583780.3614825}
\acmISBN{979-8-4007-0124-5/23/10}

\begin{document}

\title{Contrastive Representation Learning Based on Multiple Node-centered Subgraphs}



\author{Dong Li}
\affiliation{%
  \institution{Tianjin University}
  \city{Tianjin}
  \country{China}}
\email{ld2022244154@tju.edu.cn}

\author{Wenjun Wang}
\affiliation{%
  \institution{Tianjin University}
  \city{Tianjin}
  \country{China}}
\email{wjwang@tju.edu.cn}

\author{Minglai Shao}
\authornote{Corresponding author}
\affiliation{%
  \institution{Tianjin University}
  \city{Tianjin}
  \country{China}}
\email{shaoml@tju.edu.cn}

\author{Chen Zhao}
\affiliation{%
  \institution{Baylor University}
  \city{Waco, Texas }
  \country{USA}}
\email{chen_zhao@baylor.edu}

\renewcommand{\shortauthors}{Li et al.}

\begin{abstract}
    As the basic element of graph-structured data, node has been recognized as the main object of study in graph representation learning. A single node intuitively has multiple node-centered subgraphs from the whole graph (e.g., one person in a social network has multiple social circles based on his different relationships). We study this intuition under the framework of graph contrastive learning, and propose a multiple node-centered subgraphs contrastive representation learning method to learn node representation on graphs in a self-supervised way. Specifically, we carefully design a series of node-centered regional subgraphs of the central node. Then, the mutual information between different subgraphs of the same node is maximized by contrastive loss. Experiments on various real-world datasets and different downstream tasks demonstrate that our model has achieved state-of-the-art results.
\end{abstract}




\maketitle

\section{Introduction}
\label{sec:intro}
Graph representation learning has received increasing attention recently \cite{hamilton2017representation}, which aims to transform high-dimensional graph-structured data into low-dimensional dense vectorized representations. As the basic elements of graph-structured data, node representation has been the main object of graph representation learning. A comprehensive node representation can be well used for a variety of downstream tasks, such as node classification \cite{GCN} and link prediction \cite{grover2016node2vec}.

\begin{figure}[t]
\centering
\includegraphics[width=0.95\columnwidth]{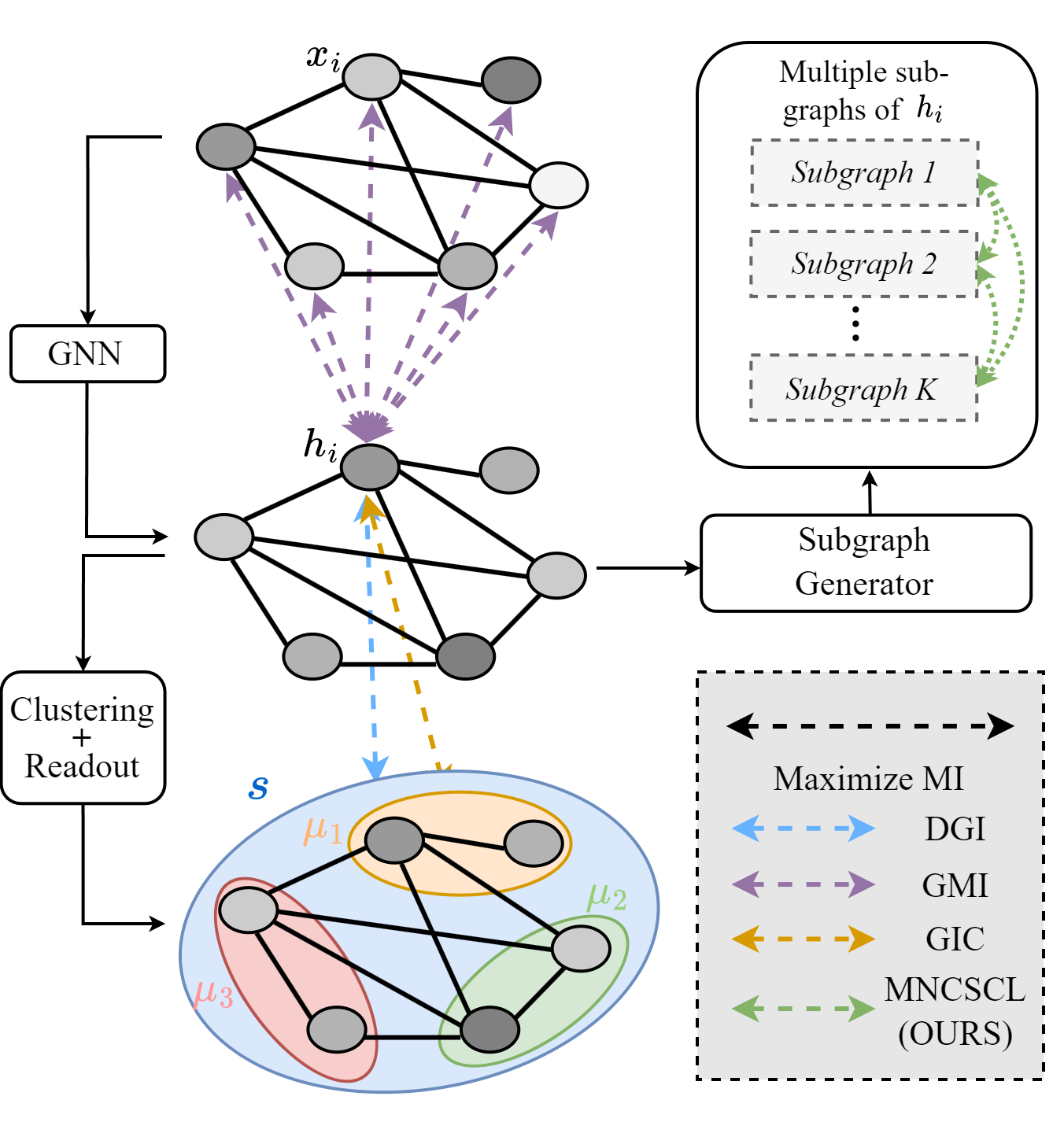} 
\caption{A descriptive illustration of different proxy tasks among DGI, GMI, GIC and our proposed MNCSCL. The two-way arrows represent MI maximization, and the different colors represent different models. The subgraph generator and multiple subgraphs of $h_i$ are described in details in Section 3.1 and Section 3.2.}
\label{fig1}
\end{figure}

A widespread graph representation learning method is the application of graph neural networks (GNN) \cite{GCN,GAT,wu2019simplifying,zhao2020multi,GraphSAGE}. But most of such methods focus on supervised learning, relying on supervised signals in the graph. For real-world networks, the acquisition of these supervised signals is often cumbersome and expensive. Self-supervised learning \cite{jing2020self} is a popular research area in recent years, which designs proxy tasks for unlabeled data to mine the representational properties of the data itself as supervised information.

\begin{figure*}[t]
\centering
\includegraphics[width=0.95\textwidth]{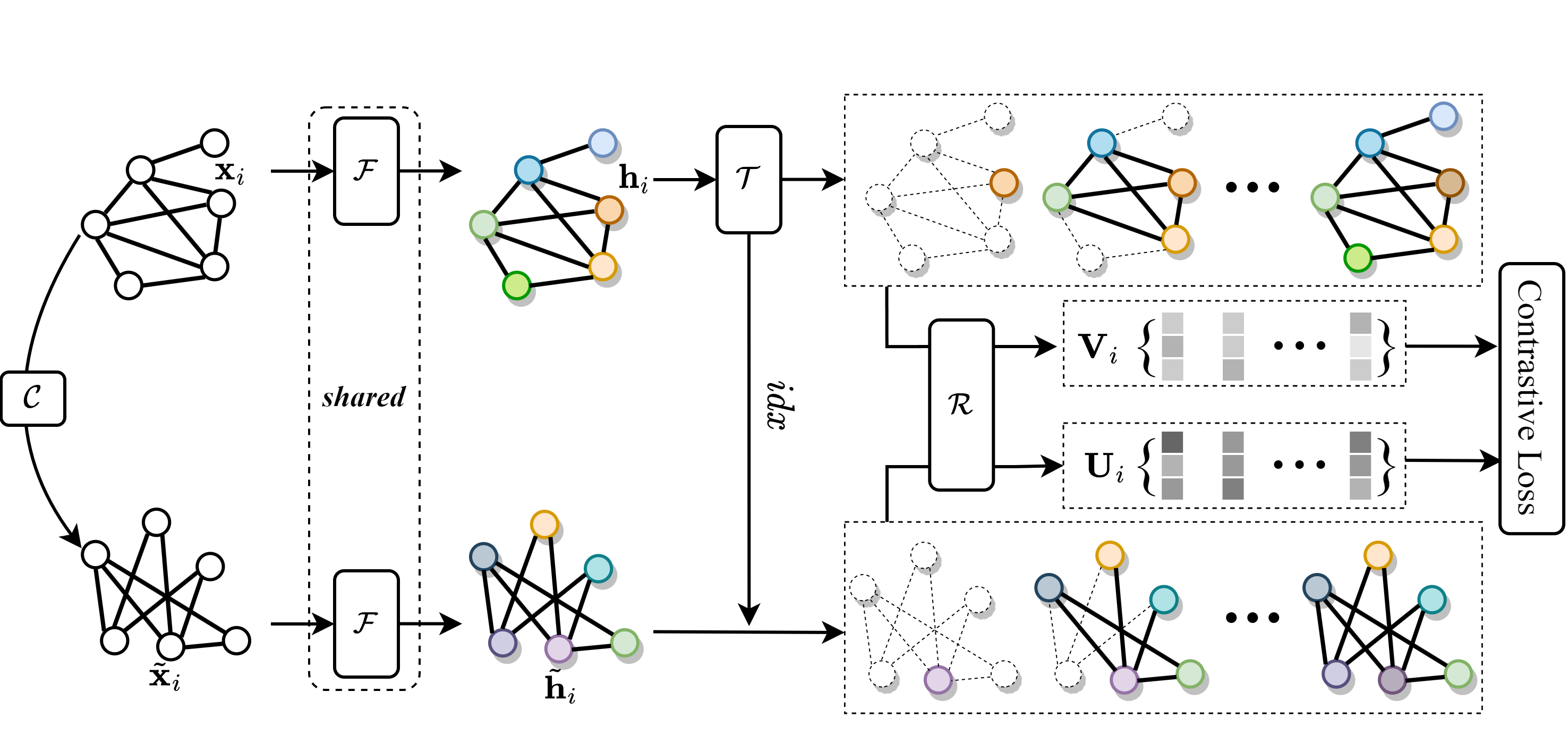} 
\caption{The pipelines of MNCSCL. For a specific node $v_i$ with attribute $\textbf{x}_i$, we first get its negative example $\tilde{\textbf{x}}_i$ through perturbing the structure and attributes of the input graph $\mathcal{G}$ with a corruption function $\mathcal{C}$. Then we use a subgraph generator $\mathcal{T}$ to get a series of node-centered subgraphs $\displaystyle\textbf{G}_i$ and their corresponding negative set $\displaystyle\tilde{\textbf{G}}_i$ from $\textbf{h}_i$ and $\tilde{\textbf{h}}_i$ (obtained by a shared encoder $\mathcal{F}$). Finally, the mutual information between $\displaystyle\textbf{G}_i$ and $\displaystyle\tilde{\textbf{G}}$ is maximized in the latent space $\displaystyle\textbf{V}_i$ and $\displaystyle\textbf{U}_i$ (obtained by a readout function $\mathcal{R}$) by contrastive loss. For more details, refer to the ``overall framework'' subsection in Section 3.}
\label{fig2}
\end{figure*}

As one of the representative methods of self-supervised learning, the proxy task of contrastive learning is to maximize the Mutual Information (MI) \cite{qiu2020gcc} between the input and related content \cite{xu2021self}. For example,  Deep Graph Infomax (DGI) \cite{DGI} maximizes MI between local view and global view of the input graph and its corresponding corrupted graph. Graphical Mutual Information (GMI) \cite{GMI} doesn’t use corruption function, instead, it maximizes the MI between the hidden representation of nodes and their original local structure. Graph InfoClust (GIC) \cite{GIC}, on the other hand, maximizes the MI between the node representation and its corresponding cluster representation on the basis of DGI. Although these methods have achieved many advances, they all focus on the MI between node embeddings and only one related graph structure, as shown in Figure \ref{fig1}.

In reality, we can look at a specific thing from multiple perspectives. For graph data, we can observe individual nodes in a graph from multiple perspectives, yet little literature has focused on this. Intuitively, for an individual in a social network, there may be a social circle of relatives based on blood relations, a social circle of colleagues based on work relations, and other social circles of friends with many different interests. If we analyze this individual from these different social circles, it is actually equivalent to learning from multiple perspectives on the nodes in this network.

Based on this intuition, we propose Multiple Node-centered Subgraphs Contrastive Representation Learning (MNCSCL). MNCSCL takes each node in the network as the center and samples its node-centered regional subgraphs under different semantics, thus forming several different perspectives of the corresponding node, as shown in Figure 1. More specifically, we first generate a negative example through the corrupt function, then generate a series of node-centered subgraphs of the original graph by the view generator, and sample the corresponding subgraphs on the negative example. Then, these subgraphs are fed into graph neural network encoders to obtain the representations of central nodes and its subgraphs after pooling. Finally, the mutual information between different subgraphs of the same node is maximized in the latent space by contrastive learning objective function. Experimental results on a variety of datasets demonstrate the superb performance of our design. The major contributions of this paper are as follows:
\begin{itemize}
\item We propose a novel framework to learn node representation through multiple node-centered subgraphs of nodes, which is a novel idea in current work to observe a single node from multiple perspectives.
\item We carefully design five node-centered subgraphs and analyze the influence of different subgraphs on the learning quality of node representation through extensive experiments, which is of reference significance.
\item We evaluated MNCSCL on five standard datasets and two downstream tasks to validate the effectiveness of the proposed method. Our experiments show that the contrastive learning of multiple subgraphs outperforms the above mentioned single-subgraph contrastive learning in terms of results.
\end{itemize}

\section{Related Work}
\label{sec:related}
\subsection{Graph Representation Learning Based on Contrastive Learning}

Inspired by the advances of contrastive learning in fields such as CV and NLP, some work has started to apply contrastive learning on graph data for graph representation learning. The objective of graph contrastive learning is to maximize the MI between similar instances in a graph \cite{wu2021self}, and its model design focuses on three modules: data augmentation, proxy tasks, and contrastive objectives. Among them, the most important modules is the proxy task, which describes the definition of similar instances (i.e. positive example) and dissimilar instances (i.e. negative example).  DGI \cite{DGI} extends the idea of DIM \cite{DIM} to graph data to learn node representations by maximizing the MI between local node representation and global graph representation. GMI \cite{GMI} takes nodes and their neighbors as objects of study and maximizes the MI between hidden representation of each node and the original features of its neighboring nodes. GIC \cite{GIC} clusters the nodes in the graph by a differentiable version of K-means clustering, and then maximizes the MI between the node representation and its corresponding cluster summaries. SUBG-CON \cite{SUBG-CON} obtains the context subgraph of each node by subgraph sampling based data augmentation, and then maximizes the consistency between them. Despite the good results achieved, these works perform graph contrastive learning only on a single perspective for nodes.

\subsection{Multi-view Contrastive Learning}

Recently, multi-view representation learning has become a rapidly growing direction in machine learning and data mining areas \cite{li2018survey, peng2022towards, zhao2019rank,zhao2020multi,zhao2021fairness,zhao2021fairnessphd}. It has had a lot of success in areas such as computer vision. For instance, Contrastive Multiview Coding (CMC) \cite{CMC} uses contrastive learning to maximize the mutual information between multiple views of a dataset to perform representation learning of images. MVGRL \cite{MVGRL} obtains multiple views of graph through data augmentation, and they find out that unlike visual representation learning, increasing the number of views of the entire graph to more than two by data augmentation does not improve performance. Unlike the multi-view graph contrastive learning summarized in MVGRL which focuses on node attributes at graph-level, our multiple node-centered subgraphs in this paper focus more on the differences in structure at node-level.

\section{Methodology}
\label{sec:methodology}
\textbf{Problem definition.} Given a graph $\mathcal{G}=(\mathcal{V},  \mathcal{E})$ with $N$ nodes, where $\displaystyle\mathcal{V}=\{v_1,v_2,...,v_N\}$ and  $\mathcal{E}$  represent the node set and the edge set respectively. $\displaystyle\textbf{X}=\{\textbf{x}_1,\textbf{x}_2,...,\textbf{x}_N\}\in\mathbb{R}^{N\times F}$ is the node features matrix, where $\displaystyle\textbf{x}_i\in\mathbb{R}^F$ denotes the features of dimension $F$ for node $v_i$. We use the adjacency matrix $\displaystyle\textbf{A}\in\mathbb{R}^{N\times N}$ to represent the connectivity of the graph, where $\textbf{A}(i,j)=1$ if nodes $v_i$ and $v_j$ are linked, and $\textbf{A}(i,j)=0$ otherwise. In this way, a  graph can also be represented as $\mathcal{G}=(\textbf{X},  \textbf{A})$. If $\displaystyle\mathcal{V}'\subset \mathcal{V}$ is a subset of vertices of $\mathcal{G}$ and $\mathcal{E}'$ consists of all of the edges in $\mathcal{E}$ that have both endpoints in $\mathcal{V}$, then the subgraph $\mathcal{S}=(\mathcal{V}', \mathcal{E}')$ of graph $\mathcal{G}$ is an induced subgraph. The subgraphs mentioned in this paper are all induced subgraphs. 

\begin{table}[t]
\caption{Summary of notations. The first five notations are described in detail in the following subsections. }
\centering
\resizebox{1\columnwidth}{!}{
\begin{tabular}{cl}
    \toprule
    \textbf{Notation} & \textbf{Meaning}\\
    \specialrule{0em}{0pt}{-0.2pt}
    \midrule
    $\mathcal{F}$ & Shared encoder\\
    \specialrule{0em}{0pt}{-0.2pt} \cmidrule(lr){1-2}
    $\mathcal{C}$ & Corruption function\\
    \specialrule{0em}{0pt}{-0.2pt} \cmidrule(lr){1-2}
    $\mathcal{T}$ & Subgraph generator \\
    \specialrule{0em}{0pt}{-0.2pt} \cmidrule(lr){1-2}
    $\mathcal{R}$ & Readout function\\
    \specialrule{0em}{0pt}{-0.2pt} \cmidrule(lr){1-2}
    $\mathcal{D}$ & Discriminator\\
    \specialrule{0em}{0pt}{-0.2pt} \cmidrule(lr){1-2}
    $N$ & Number of nodes in the input graph\\
    \specialrule{0em}{0pt}{-0.2pt} \cmidrule(lr){1-2}
    $F$ & Feature dimension of each node \\
    \specialrule{0em}{0pt}{-0.2pt} \cmidrule(lr){1-2}
    $F'$ & Feature dimension of each node representation \\
    \specialrule{0em}{0pt}{-0.2pt} \cmidrule(lr){1-2}
    $K$ & Number of node-centered subgraphs sampled by $\mathcal{T}$ \\
    \specialrule{0em}{0pt}{-0.2pt} \cmidrule(lr){1-2}
    $N'$ & \makecell[l]{Number of nodes in the corresponding \\ node-centered subgraph} \\
    \specialrule{0em}{0pt}{-0.2pt} \cmidrule(lr){1-2}
    $d$ & Range of neighbors in the \textbf{neighboring subgraph} \\
    \specialrule{0em}{0pt}{-0.2pt} \cmidrule(lr){1-2}
    $l$ & \makecell[l]{Number of nodes that are most similar to the \\ central node for \textbf{intimate subgraph}} \\
    \specialrule{0em}{0pt}{-0.2pt} \cmidrule(lr){1-2}
    $C$ & number of clusters for \textbf{communal subgraph} \\
    \specialrule{0em}{0pt}{-0.2pt} \cmidrule(lr){1-2}
    $\eta$ & self-weighted factor for \textbf{full subgraph} \\
    \specialrule{0em}{0pt}{-0.2pt} \cmidrule(lr){1-2}
    $\mathcal{G}, \tilde{\mathcal{G}}$ & Input graph and its negative example obtained by $\mathcal{C}$\\
    \specialrule{0em}{0pt}{-0.2pt} \cmidrule(lr){1-2}
    $\mathcal{G}_i^k, \tilde{\mathcal{G}}_i^k$ & \makecell[l]{The $k$-th node-centered subgraph of node $v_i$ and \\ its negative example}\\
    \specialrule{0em}{0pt}{-0.2pt} \cmidrule(lr){1-2}
    $\textbf{H}_i^k, \tilde{\textbf{H}}_i^k$ & \makecell[l]{Node representation matrix of the $k$-th node-centered \\ subgraph of node $v_i$ and its negative example}\\
    \specialrule{0em}{0pt}{-0.2pt} \cmidrule(lr){1-2}
    $\textbf{A}_i^k, \tilde{\textbf{A}}_i^k$ & \makecell[l]{Adjacency matrix of the $k$-th node-centered subgraph \\ of node $v_i$ and its negative example}\\
    \specialrule{0em}{0pt}{-0.2pt} \cmidrule(lr){1-2}
    $\textbf{v}_i^k, \textbf{u}_i^k$ & \makecell[l]{Representation of the $k$-th node-centered subgraph of \\ node $v_i$ and its negative example, obtained by $\mathcal{R}$}\\
    \bottomrule 
\end{tabular}}
\label{table1}
\end{table}

The goal of self-supervised graph representation learning is to learn a encoder $\displaystyle \mathcal{F}:\mathbb{R}^{N\times F}\times\mathbb{R}^{N\times N}\to \mathbb{R}^{N\times F'}$, which takes the features matrix $\textbf{X}$ and the adjacency matrix $\textbf{A}$ as input to get the node representation $\textbf{H}=\{\textbf{h}_1,\textbf{h}_2,...,\textbf{h}_N\}\in\mathbb{R}^{N\times F'}$ without label information, formulated as $\textbf{H}=\mathcal{F}(\textbf{X}, \textbf{A})$. The learned node representation $\textbf{H}$ can be used directly for downstream tasks such as node classification and link prediction. For the sake of clarity, we list all important notations in Table \ref{table1}.

\textbf{Overall framework.} Inspired by recent graph representation learning work based on contrastive learning, we propose MNCSCL algorithm for graph representation learning by maximizing MI of multiple node-centered subgraphs of nodes. As illustrated in Figure \ref{fig2}, if there is only a single graph provided as input, the summarized steps of MNCSCL are as follows:
\begin{itemize}
    \item Utilize a corruption function $\mathcal{C}$ to perturb the structure and attributes of the input graph $\mathcal{G}$ to obtain a negative example $\displaystyle\tilde{\mathcal{G}}=(\tilde{\textbf{X}}, \tilde{\textbf{A}})\sim\mathcal{C}(\textbf{X}, \textbf{A})$.
    
    \item Pass input graph $\mathcal{G}$ and negative example $\tilde{\mathcal{G}}$ into a shared encoder $\mathcal{F}$ to get node representation $\textbf{H}$ and $\tilde{\textbf{H}}$.
    
    \item Use a subgraph generator $\mathcal{T}$ to sample a series of node-centered subgraphs $\displaystyle\textbf{G}_i=\{\mathcal{G}_i^1,\mathcal{G}_i^2,...,\mathcal{G}_i^K\}$ for node $v_i$ from the input graph $\mathcal{G}$, where $K$ is the number of different subgraphs and $\mathcal{G}_i^k=(\textbf{H}_i^k,\textbf{A}_i^k)$, $k=1,2,...,K$. The corresponding node-centered subgraphs set $\displaystyle\tilde{\textbf{G}}_i=\{\tilde{\mathcal{G}}_i^1,\tilde{\mathcal{G}}_i^2,...,\tilde{\mathcal{G}}_i^K\}$ from the negative example $\tilde{\mathcal{G}}$ are further obtained according to $\textbf{G}_i$.
    
    \item Summarize all subgraphs in $\textbf{G}_i$ and $\tilde{\textbf{G}}_i$ through a readout function $\mathcal{R}$ to get their representations $\displaystyle\textbf{V}_i=\{\textbf{v}_i^1,\textbf{v}_i^2,...,\textbf{v}_i^K\}$ of node $v_i$ and the corresponding negative samples $\displaystyle\textbf{U}_i=\{\textbf{u}_i^1,\textbf{u}_i^2,...,\textbf{u}_i^K\}$. As an example, $\textbf{v}_i^k=\mathcal{R}(\textbf{H}_i^k)$.
    
    \item Update parameters of $\mathcal{F}$, $\mathcal{R}$ and $\mathcal{D}$ (mentioned later) by applying gradient descent to maximize Eq. (\ref{eq13}) or Eq. (\ref{eq14}).

\end{itemize}

In the following sections, we will elaborate on the crucial components mentioned above.

\subsection{Subgraph Generator}

We can deal with a whole graph from different views \cite{MVGRL}, and the same is true for any node in the graph. For a single node, there are many subgraphs centered on it (e.g., ego network \cite{zimmermann2008predicting}). If these node-centered regional subgraphs have certain semantic information (e.g., ego network represents a specific individual and other persons who have a social relationship with him), then we can treat them as different perspectives of the central node. 

The main role of the subgraph generator $\mathcal{T}$ is to sample these node-centered regional subgraphs from $\mathcal{G}$ and generate the corresponding negative samples from $\tilde{\mathcal{G}}$. Specifically, for a specific node $v_i$ and a prepared node-centered subgraph type $k$, the subgraph generator $\mathcal{T}$ first gets $idx$, a set which represents the index of chosen nodes for subgraph $\mathcal{G}_i^k$ to be obtained. Then, the node representation matrix $\textbf{H}_i^k\in\mathbb{R}^{N'\times F'}$ and adjacency matrix $\textbf{A}_i^k\in\mathbb{R}^{N'\times N'}$ of $\mathcal{G}_i^k$ are denoted respectively as
\begin{equation}
\textbf{H}_i^k=\textbf{H}_{idx,:},\  \textbf{A}_i^k=\textbf{A}_{idx,idx}, \label{eq1}
\end{equation}
where $\cdot_{idx}$ is an indexing operation and $N'$ is the length of $idx$. 

In this way, we can obtain the $k$-th node-centered subgraph $\displaystyle\mathcal{G}_i^k=(\textbf{H}_i^k,\textbf{A}_i^k)\sim\mathcal{T}(\textbf{H}, \textbf{A})$  for any specific node $v_i$. Likewise, the corresponding negative example can be obtained by $\displaystyle\tilde{\mathcal{G}}_i^k=(\tilde{\textbf{H}}_i^k,\tilde{\textbf{A}}_i^k)=(\tilde{\textbf{H}}_{idx,:}, \tilde{\textbf{A}}_{idx,idx})$.

\begin{figure}[t]
\centering
\includegraphics[width=0.95\columnwidth]{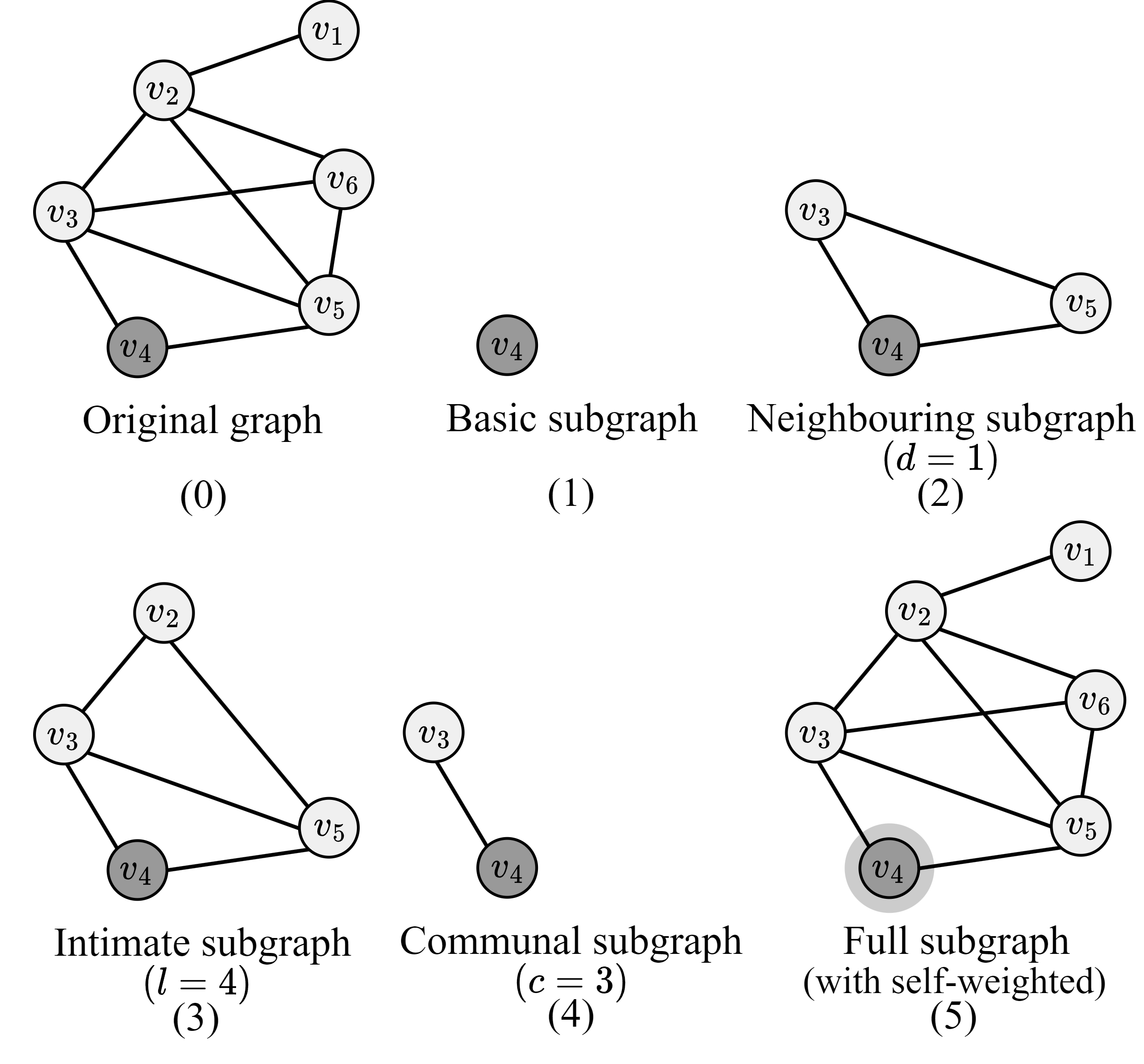} 
\caption{Five types of node-centered subgraphs for node $v_4$ from original graph. Note that nodes are divided into 3 clusters in (4), which are $\{v_1,v_2\}$, $\{v_3,v_4\}$ and $\{v_5,v_6\}$.}
\label{fig3}
\end{figure}

\subsection{Node-centered Subgraphs Design}

To learn a more comprehensive representation, we carefully design five different node-centered subgraphs, as illustrated in Figure \ref{fig3}. The details of them are as follows:

\textbf{Subgraph 1: Basic subgraph.} The basic subgraph only contains the central node itself (i.e., $N'=1$), that means for each node $v_i$:
\begin{equation}
idx=\{i\}. \label{eq2}
\end{equation}
Further, we can get the basic subgraph representation and its corresponding negative example by $\textbf{v}_i^1=\textbf{h}_i$ and $\textbf{u}_i^1=\tilde{\textbf{h}}_i$. For any specific node, basic subgraph is the purest ``subgraph'' as well as the main subgraph, which contains the most concentrated features of the node iteself.

\textbf{Subgraph 2: Neighboring subgraph.} The neighbors of the central node are often closely related to the node in structure, and study with them can better capture the structural features of node \cite{GraphSAGE}. The neighboring subgraph contains all nodes with a distance less than or equal to $d$ from the central node $v_i$, denoted as
\begin{equation}
idx=\{j|dis(v_i,v_j)\leq d\}, \label{eq3}
\end{equation}
where $dis(\cdot,\cdot)$ is a function used to get the distance between two nodes. After obtaining the neighboring subgraph $\mathcal{G}_i^2$ by Eq. (\ref{eq1}), a readout function $\displaystyle \mathcal{R}:\mathbb{R}^{N\times F'}\to \mathbb{R}^{F'}$ is used to obtain the neighboring subgraph representation and its corresponding negative example:
\begin{equation}
\textbf{v}_i^2=\mathcal{R}(\textbf{H}_i^2),\  
\textbf{u}_i^2=\mathcal{R}(\tilde{\textbf{H}}_i^2).\label{eq4}
\end{equation}

It is beneficial to learn a more comprehensive representation when we take a larger range of neighbors. But at the same time, the features of the central node will be weakened, which will cause the model to be more inclined to learn the representation of a region or even the whole graph. It follows that it's important to choose the value of $d$. As shown in the Figure \ref{fig5}, the model reaches the best performance when $d=1$ for the neighboring subgraph.

\textbf{Subgraph 3: Intimate subgraph.} The intimate subgraph takes into account the similarity that actually exists between two nodes in the input graph $\mathcal{G}$. It contains the first $l$ nodes that are most similar to the central node. This is equivalent to identifying the nodes that are structurally close to the central node from another perspective, and thus learning the structural features of the nodes more comprehensively.

The similarity between nodes is usually measured by a similarity scores matrix $\textbf{S}\in\mathbb{R}^{N\times N}$, where $\textbf{S}(i,j)$ measures the similarity between nodes $v_i$ and $v_j$. Here we follow the personalized pagerank (PPR) algorithm \cite{jeh2003scaling} as introduced in \cite{zhang2020graph}. The similarity scores matrix $\textbf{S}$ based on the PPR algorithm can be denoted as
\begin{equation}
\textbf{S}=\alpha\cdot (\textbf{I}-(1-\alpha)\cdot \bar{\textbf{A}})^{-1},\label{eq5}
\end{equation}
where \textbf{I} is the identity matrix and $\bar{\textbf{A}}=\textbf{A}\textbf{D}^{-1}$ denotes the colum-normalized adjacency matrix. $\textbf{D}$ is the diagonal matrix corresponding to $\textbf{A}$ with $\textbf{D}(i,i)=\sum_j\textbf{A}(i,j)$ on its diagonal.  $\alpha\in [0,1]$ is a parameter
which is set as 0.15 in \cite{SUBG-CON}. For a specific node $v_i$, the subgraph generator $\mathcal{T}$ chooses its top-$l$ similar nodes (i.e., $N'=l$) to generate intimate subgraph with $\textbf{S}(i,:)$, which can be denoted as
\begin{equation}
idx=top\_rank(\textbf{S}(i,:), l),\label{eq6}
\end{equation}
where $top\_rank(\cdot)$ is a function that selects the top-$l$ values from a vector and return the corresponding indices. Same as \textit{Subgraph 2} subsection, we can obtain the intimate subgraph representation and its corresponding negative example with $\textbf{v}_i^3=\mathcal{R}(\textbf{H}_i^3)$ and $\textbf{u}_i^3=\mathcal{R}(\tilde{\textbf{H}}_i^3)$.

\textbf{Subgraph 4: Communal subgraph.} In graph clustering, nodes in a uniform cluster tend to have similarity in attributes. Therefore, we can select all nodes in the cluster to which the central node belongs to get the communal subgraph. These nodes with similar attributes to the central node can help the model better learn the attribute features of the central node. 

It is particularly noteworthy that the other subgraphs are obtained independently of the node attributes (i.e., $\textbf{H}$), but the communal subgraph is attribute-related. Since $\textbf{H}$ will change during training, a fixed $idx$ before the model training as other subgraphs may lead to undesirable results.

There are already many methods to perform graph clustering \cite{schaeffer2007graph}. We tried three different clustering strategies based on $K$-means clustering \cite{macqueen1967classification} due to the stable and excellent performance of it.

\begin{itemize}
    \item \textbf{Strategy 1: Precomputed K-means.} Same as other subgraphs, the node-centered subgraphs are sampled before model training starts. Specifically, the traditional $K$-means clustering algorithm is used to cluster the nodes in the input graph $\mathcal{G}$. For any specific node $v_i$, take all the indices of nodes in its cluster as $idx$ and further compute $\textbf{v}_i^4$ and $\textbf{u}_i^4$.
    \item \textbf{Strategy 2: A differentiable version of K-means.} To update the communal subgraph during training, we need an end-to-end clustering algorithm. Here we follow a differentiable version of K-means as introduced in \cite{wilder2019end}. 
    For each node $v_i$, let $\mathbf{\mu}_c$ denote the center of cluster $c$ and 
    $\hat{\mathbf{\gamma}}_{ic}$ (s.t., $\sum_c\hat{\mathbf{\gamma}}_{ic}=1, \forall i$) 
    denotes the degree to which node $v_i$ is assigned to cluster $c$. 
    Suppose that the number of clusters to be obtained is $C$, ClusterNet updates $\mathbf{\mu}_c$ via an iterative process by alternately setting.
    \begin{equation}
    \mathbf{\mu}_c=\frac{\sum_i\hat{\mathbf{\gamma}}_{ic}\textbf{h}_i}{\sum_i\hat{\mathbf{\gamma}}_{ic}}\
    \ c=1,...,C,\label{eq7}
    \end{equation}
    and
    \begin{equation}
    \hat{\mathbf{\gamma}}_{ic}=\frac{exp(-\beta\cdot sim(\textbf{h}_i,\mathbf{\mu}_c))}{\sum_cexp(-\beta\cdot sim(\textbf{h}_i,\mathbf{\mu}_c))}\
    \ c=1,...,C,\label{eq8}
    \end{equation}
    where $\beta$ is an inverse-temperature hyperparameter, the standard K-means assignment is recovered when $\beta\to\infty$. $sim(\cdot,\cdot)$ denotes a similarity function between two instances. Eventually, we can get the communal subgraph representation by
    \begin{equation}
    \textbf{v}_i^4=\sigma\left(\sum_{c=1}^C\hat{\mathbf{\gamma}}_{ic}\mathbf{\mu}_c\right),\label{eq9}
    \end{equation}
    where $\sigma(\cdot)$ is the logistic sigmoid nonlinearity. Since this method does not get $idx$, we simply get negative example $\displaystyle\{\textbf{u}_1^4,\textbf{u}_2^4,...,\textbf{u}_N^4\}$ of all nodes by row-wise shuffling of $\displaystyle\{\textbf{v}_1^4,\textbf{v}_2^4,...,\textbf{v}_N^4\}$.

    \item \textbf{Strategy 3: An end-to-end version of K-means with an estimation network.} In this way, we replace the iterative process in \textit{Strategy 2} with an estimation network, which utilizes a multi-layer neural network to directly predict the degree to which each node belongs to each cluster $\displaystyle\hat{\mathbf{\gamma}}\in\mathbb{R}^{N\times C}$, denoted as
    \begin{equation}
    \hat{\mathbf{\gamma}}=softmax(MLP(\textbf{H};\theta)),\label{eq10}
    \end{equation}
    where $softmax(\cdot)$ is a softmax nonlinearity and $MLP(\cdot)$ is a multi-layer neural network with trainable parameters $\theta$. Then we get the communal subgraph representation by Eq. (\ref{eq8}) and Eq. (\ref{eq9}) as same as \textit{Strategy 2}.
    
\end{itemize}

The comparison of the three strategies is shown in Figure \ref{fig5}. After weighing both accuracy and efficiency, we chose \textit{Strategy 2} to generate the communal subgraph.

\begin{figure}[t]
\centering
\includegraphics[width=1.0\columnwidth]{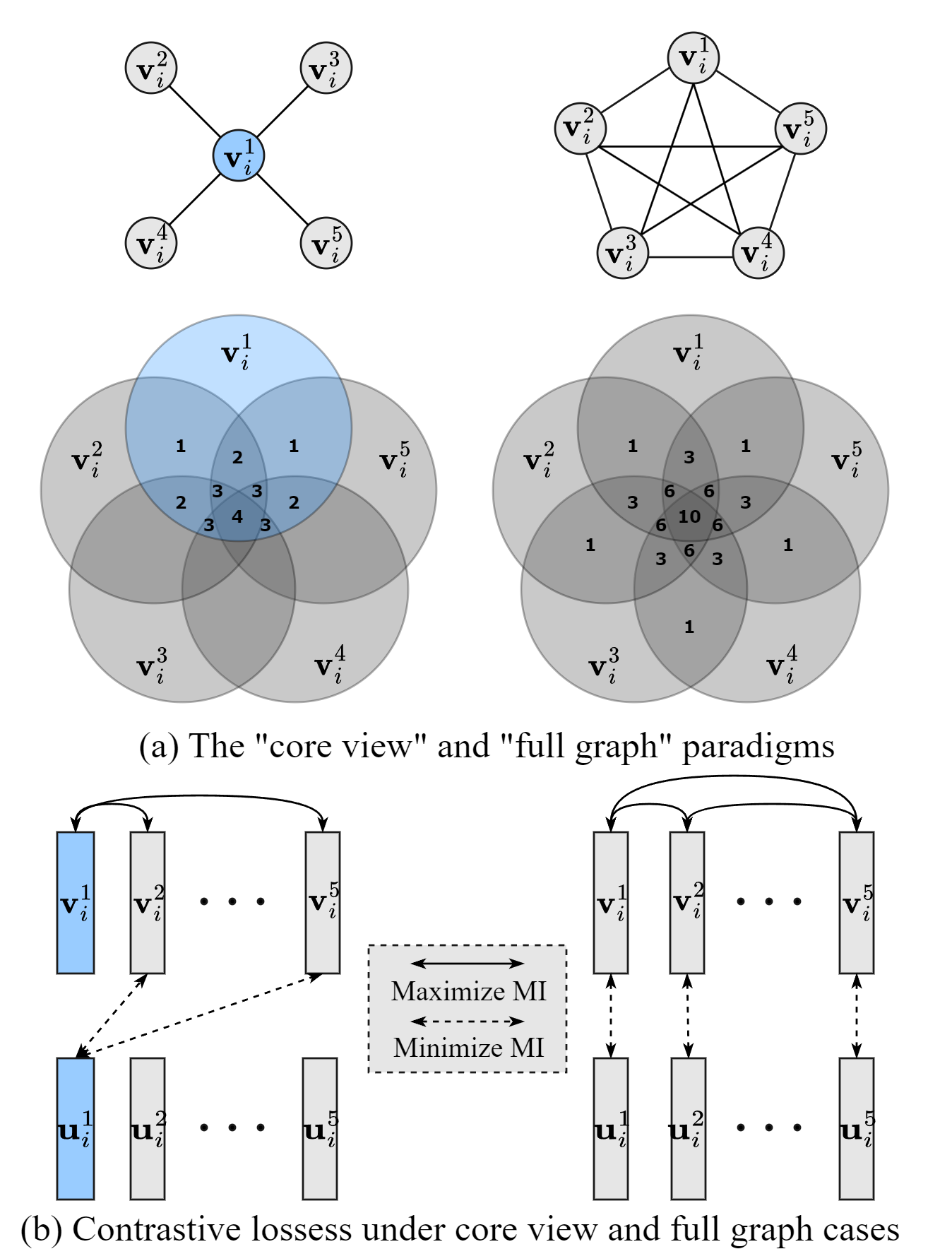} 
\caption{Take the five node-centered subgraphs of node $v_i$ as an example. (a) The ``core view'' (left) and ``full graph'' (right) paradigms. The numbers within the regions represent the number of MI in this region. For example, if we select all 5 node-centered subgraphs under the full graph case, MI will be calculated once between every two subgraphs, and hence ismarked with the number 10. (b) Contrastive lossess under core view and full graph cases. Refer to Section 3.3 for more details.}
\label{fig4}
\end{figure}

\begin{table*}[t]
\caption{The statistics of all the five datasets. *Note that the node classification on PPI dataset is a multilabel classification problem.}
\centering

\begin{tabular}{cccccccc}
\toprule
\textbf{Task} & \textbf{Dataset} & \textbf{Type} & \textbf{\#Nodes} & \textbf{\#Edges} & \textbf{\#Features} & \textbf{\#Classes} & \textbf{Train / Val / Test} \\

\specialrule{0em}{0pt}{-0.2pt}
\midrule

\multirow{3}{*}{\makecell[c]{Node classification\\ \& Link prediction\\(Transductive)}} & Cora & Citation network & 2,708 & 5,429 & 1,433 & 7 & 0.05 / 0.18 / 0.37\\
 & Citeseer & Citation network & 3,327 & 4,732 & 3,703 & 6 & 0.04 / 0.15 / 0.30\\
 & Pubmed & Citation network & 19,717 & 44,338 & 500 & 3 & 0.003 / 0.03 / 0.05\\

\hline
\specialrule{0em}{1pt}{1pt}

\multirow{2}{*}{\makecell[c]{Node classification\\(Inductive)}} & Reddit & Social network & 232,965 & 11,606,919 & 602 & 41 & 0.66 / 0.10 / 0.24\\
 & PPI & Protein network & 56,944 & 818,716 & 50 & 121* & 0.79 / 0.11 / 0.10\\

\specialrule{0em}{0pt}{-0.2pt}
\bottomrule 
\end{tabular}
\label{table2}
\end{table*}

\textbf{Subgraph 5: Full subgraph.} To learn a comprehensive representation of a node, it is essential to observe it from a global perspective. The full subgraph contains all the nodes (i.e. $N'=N$) in the input graph $\mathcal{G}$, e.g., for any specific node $v_i$,
\begin{equation}
idx=\{j|j=1,2,...,N\}.\label{eq11}
\end{equation}

Compared with the previous subgraphs, the full subgraph contains far more nodes than they do, which extremely weakens the specificity of the central node. At the same time, it is not conducive to learning a specialized node representation as all nodes have the same full subgraph. Based on these ideas, we propose full subgraph for specific node $v_i$ with self-weighted, denoted as
\begin{equation}
\label{eq12}
\textbf{v}_i^5=(1-\eta)\mathcal{R}(\textbf{H}_i^5)+\eta\textbf{h}_i,
\end{equation}
where $\eta\in[0,1]$ is a self-weighted factor.

So far, we have introduced five carefully designed node-centered subgraphs. It is noted that subgraphs other than \textit{Subgraph 4} can be precomputed before model training starts, which allows us to quickly obtain these subgraphs during training by performing only one calculation before training.

\subsection{Contrastive Loss}

The key idea of self-supervised comparative learning is to define a proxy task to generate positive and negative samples. The encoder $\mathcal{F}$ that generates the node representation is trained by contrast between positive and negative samples. 

To handle multiple subgraphs we obtained before, we take the ``core view'' (CV) and ``full graph'' (FG) paradigms as introduced in CMC \cite{CMC}, as shown in Figure \ref{fig4}.(a). Further, the contrastive lossess under core view and full graph cases are illustrated in Figure \ref{fig4}.(b). Specifically, in the core view case, we regard \textit{Subgraph 1} as the most critical node-centered subgraph. It and its corresponding negative sample constitute positive and negative pairs with \textit{Subgraph 2$\thicksim$5}, respectively. As for the full graph case, any two between \textit{Subgraph 1$\thicksim$5} constitute positive pairs, and \textit{Subgraph 1$\thicksim$5} respectively form negative pairs with their corresponding negative examples.

After defining the proxy task, we follow the intuitions from DGI \cite{DGI} and use a noise-contrastive type objective with a standard binary cross-entropy (BCE) loss between positive examples and negative examples. MNCSCL's objective under core view case is
\begin{equation}
\label{eq13}
    \begin{split}
        \mathcal{L}_{CV}=\sum_{i=1}^N\sum_{j=2}^K\mathbb{E}_{(\textbf{X},\textbf{A})}\left[log\mathcal{D}(\textbf{v}_i^1,\textbf{v}_i^j)\right]\\
        +\sum_{i=1}^N\sum_{j=2}^K\mathbb{E}_{(\tilde{\textbf{X}},\tilde{\textbf{A}})}\left[log(1-\mathcal{D}(\textbf{u}_i^1,\textbf{v}_i^j))\right],
    \end{split}
\end{equation}
where $K$ is the number of selected node-centered subgraphs and $\mathcal{D}:\mathbb{R}^{F'}\times\mathbb{R}^{F'}\to\mathbb{R}$ is a discriminator which is used for estimating the MI by assigning higher scores to positive examples than negatives. When using the full graph case, the objective becomes
\begin{equation}
\label{eq14}
    \begin{split}
        \mathcal{L}_{FG}=\sum_{i=1}^N\sum_{j=1}^{K-1}\sum_{k=j+1}^K\mathbb{E}_{(\textbf{X},\textbf{A})}\left[log\mathcal{D}(\textbf{v}_i^j,\textbf{v}_i^k)\right]\\
        +\sum_{i=1}^N\sum_{j=1}^K\mathbb{E}_{(\tilde{\textbf{X}},\tilde{\textbf{A}})}\left[log(1-\mathcal{D}(\textbf{v}_i^j,\textbf{u}_i^j))\right].
    \end{split}
\end{equation}

The Eq. (\ref{eq13}) and Eq. (\ref{eq14}) are used as the contrastive loss  in the experiments respectively.

\section{Experiments}
\label{sec:experiments}
\begin{table*}[t]
\caption{The classification accuracy (in \%) on the transductive datasets and the micro-averaged F1 ($\times 100$) on the inductive datasets of the node classification task. Some results are directly taken from their original papers (DGI, GMI\#inductive, GIC and GRACE\#inductive), and other compared results are taken from \cite{SUBG-CON,SUGRL}. The second column is the data used in the training process (\textbf{X}: features matrix, \textbf{A}: adjacency matrix, \textbf{Y}: labels). The best result for each dataset is indicated by bolded.}
\centering

\begin{tabular}{lcccccccc}
\toprule
\multirow{2}{*}{\textbf{Method}} & \multicolumn{3}{c}{\textbf{Input}} & \multicolumn{3}{c}{\textbf{Transductive}}  & \multicolumn{2}{c}{\textbf{Inductive}}\\
\specialrule{0em}{0pt}{-0.5pt}
\cmidrule(r){2-4} \cmidrule(r){5-7} \cmidrule(r){8-9}

& $\textbf{X}$ & $\textbf{A}$ & $\textbf{Y}$ & \textbf{Cora} & \textbf{Citeseer} & \textbf{Pubmed} & \textbf{Reddit} & \textbf{PPI}\\
\specialrule{0em}{0pt}{-0.2pt}
\midrule
Raw features & \checkmark &  &  &56.6 ± 0.4 & 57.8 ± 0.2 & 69.1 ± 0.2 & 58.5 ± 0.1 & 42.5 ± 0.3\\
Deep Walk &  & \checkmark &  & 67.2 & 43.2 & 65.3 & 32.4 & 52.9\\
\hline
\specialrule{0em}{1pt}{1pt}
GCN & \checkmark & \checkmark & \checkmark & 81.4 ± 0.6 & 70.3 ± 0.7 & 76.8 ± 0.6 & 93.3 ± 0.1 & 51.5 ± 0.6\\
FastGCN & \checkmark & \checkmark & \checkmark & 78.0 ± 2.1 & 63.5 ± 1.8 & 74.4 ± 0.8 & 89.5 ± 1.2  & 63.7 ± 0.6 \\
\hline
\specialrule{0em}{1pt}{1pt}
DGI & \checkmark & \checkmark &  & 82.3 ± 0.6 & 71.8 ± 0.7 & 76.8 ± 0.6 & 94.0 ± 0.1 & 63.8 ± 0.2\\
GMI & \checkmark & \checkmark &  & 83.0 ± 0.2 & 72.4 ± 0.2 & 79.9 ± 0.4 & 95.0 ± 0.02  & 65.0 ± 0.02\\
GIC & \checkmark & \checkmark &  & 81.7 ± 0.8 & 71.9 ± 0.9 & 77.4 ± 0.5 & - & - \\
GRACE & \checkmark & \checkmark &  & 83.1 ± 0.2 & 72.1 ± 0.1 & 79.6 ± 0.5 & 94.2±0.0 & 66.2±0.1\\
MVGRL & \checkmark & \checkmark &  &  82.9 ± 0.3 & 72.6 ± 0.4 & 80.1 ± 0.7 & - & - \\
\hline
\specialrule{0em}{1pt}{1pt}
MNCSCL-FG & \checkmark & \checkmark &  & 84.3 ± 0.5  & 73.2 ± 0.6 & 80.0 ± 0.4 & 95.2 ± 0.1 & \textbf{67.3 ± 0.2}\\
MNCSCL-CV & \checkmark & \checkmark &  & \textbf{84.7 ± 0.3} & \textbf{73.8 ± 0.5} & \textbf{81.5 ± 0.4} & \textbf{95.8 ± 0.1} & 67.1 ± 0.2\\
\specialrule{0em}{0pt}{-0.2pt}
\bottomrule 
\end{tabular}
\label{table3}
\end{table*}

\subsection{Experimental Settings}

\textbf{Datasets.} We use 5 commonly used benchmark datasets in the previous work \cite{GraphSAGE,GIC} for node classification and link prediction downstream tasks, including 3 transductive citation networks (i.e., Cora, Citeseer, and Pubmed), a inductive large social network (i.e., Reddit) and a inductive protein-protein interaction dataset that contains multiple graphs with multiple labels (i.e., PPI). 
\begin{itemize}
    \item \textbf{Cora, Citeseer, and PubMed} are all citation networks, with Cora and Citeseer focusing on papers in computer science and information science, and Pubmed containing a large amount of literature information in medical and life science fields. They represent citation relationships between papers through graph data structure, where each node represents a paper and the edges represent the citation relationships between papers.
    \item \textbf{Reddit} is a collection of information from Reddit, the world's largest social news aggregation, discussion and community site. the Reddit dataset provides a large amount of user-generated content, including posts, comments, polls and more. In Reddit, posts are represented as nodes, and the connections between them correspond to user comments.
    \item \textbf{PPI} is a protein-protein interaction dataset that contains multiple graphs with multiple labels. PPI contains the interaction relationships between proteins that can form a network or graph structure. Each node represents a protein, while edges indicate interactions between proteins. 
\end{itemize}
\noindent We use all five datasets in the node classification task and follow the settings in the division of the training set and the test set as same as \cite{DGI}. In the link prediction task, we use Cora, Citeseer, and Pubmed datasets and follow the setup described in \cite{kipf2016variational}. The statistics of all the datasets are shown in Table \ref{table2}.


\textbf{Baseline methods.} In node classification task, the compared methods include direct use of row features, 1 traditional unsupervised algorithm (i.e., Deep Walk \cite{deepwalk}), two supervised graph neural networks (i.e., GCN \cite{GCN} and FastGCN \cite{FastGCN}) and 5 state-of-the-art self-supervised methods(i.e., DGI, GMI, GIC, GRACE \cite{GRACE} and MVGRL \cite{MVGRL}). 
\begin{itemize}
    \item \textbf{DGI} is one of the classical methods of graph representation learning based on contrastive learning. It aims to maximize the MI between the local perspective and the global perspective of the input graph, as well as the corresponding corrupted graph.
    \item \textbf{GMI} draws on the ideas of DGI, but rather than employing a corruption function, this approach focuses on maximizing the MI between the hidden representations of nodes and their original local structure.
    \item \textbf{GIC} is also inspired by DGI, its objective is to maximize the MI between the node's representation and the representation of the cluster to which it is assigned.
    \item \textbf{GRACE} propose a novel framework for unsupervised graph representation learning by leveraging a contrastive objective at the node level. In order to enhance the contrast effect, they created two sets of negative pairs, one within the same view and the other across different views.
    \item \textbf{MVGRL} performs self-supervised learning by contrasting structural views of graphs, where they contrast first-order neighbors of nodes as well as a graph diffusion, with good results.
\end{itemize}

In link prediction, we directly follow the effective link prediction methods used in GIC (i.e., Deep Walk, Spectral Clustering (SC) \cite{tang2011leveraging}, VGAE \cite{kipf2016variational}, ARGVA \cite{pan2019learning}, DGI and GIC).
\begin{itemize}
    \item \textbf{Spectral Clustering} is initially introduced to solve the node partitioning problem in graph analysis. It has demonstrated satisfactory performance across diverse domains, such as graphs, text, images, and microarray data. Its effectiveness has been widely acknowledged in these areas.
    \item \textbf{VGAE} is an unsupervised learning framework designed for graph-structured data, utilizing the variational auto-encoder (VAE) methodology. In VGAE, a GNN-based encoder is employed to generate node embeddings, while a straightforward decoder is used to reconstruct the adjacency matrix.
    \item \textbf{ARGVA} is a graph embedding framework specifically designed for graph data, incorporating adversarial learning techniques. Similar to VGAE, ARGVA adopts a similar structure, but it learns the underlying data distribution through an adversarial approach.
\end{itemize}

\textbf{Evaluation metrics.} For the node classification task, we classify the test set by a logistic regression classifier, and then evaluate the performance using classification accuracy for transductive datasets (i.e., Cora, Citeseer and Pubmed) and micro-averaged F1 score for inductive datasets (i.e., Reddit and PPI). Suppose that TP, FN, FP and TN represent the number of true positives, false negatives, false positives, and true negatives,
respectively. Then classification accuracy can be calculated by $accuracy=(TP + TN) / (TP + FP + TN + FN)$. Also micro-averaged F1 score can be calculated by $F1-Score = 2*precision*recall/(precision+recall)$, where $precision= TP / (TP + FP)$ and $recall = TP / (TP + FN)$. For the link prediction task, we use the AUC score (the area under ROC curve) and the AP score (the area under Precision-Recall curve) for evaluation. The closer the AUC score and the AP score approaches 1, the better the performance of the algorithm is.

\textbf{Training strategy.} We implement MNCSCL using PyTorch \cite{ketkar2021introduction} on 4 NVIDIA GeForce RTX 3090 GPUs and use Adam optimizer \cite{kingma2014adam} with an initial learning rate of 0.001 (specially, 0.0001 for Reddit) during training. We follow settings in DGI that using an early stopping strategy with a patience of 20 epochs for transductive datasets and a fixed number of epochs (150 on Reddit, 20 on PPI) for inductive datasets. For large graphs, we adopt the sampling strategy used by GraphSAGE \cite{GraphSAGE}.

\begin{table*}[t]
\caption{The AUC scores (in \%) of the link prediction task. The results of the compared methods are replicated from \cite{GIC}. The best result for each dataset is indicated by bolded.}
\centering
\begin{tabular}{lcccccc}
\toprule
\multirow{2}{*}{\textbf{Method}} & \multicolumn{2}{c}{\textbf{Cora}} & \multicolumn{2}{c}{\textbf{Citeseer}} & \multicolumn{2}{c}{\textbf{Pubmed}} \\
\cmidrule(r){2-3} \cmidrule(r){4-5} \cmidrule(r){6-7}
& AUC & AP & AUC & AP & AUC & AP \\
\midrule

DeepWalk & 83.1 ± 0.01 & 85.0 ± 0.00 & 80.5 ± 0.02 & 83.6 ± 0.01 & 84.4 ± 0.00 & 84.1 ± 0.00 \\
Spectral Clustering & 84.6 ± 0.01 & 88.5 ± 0.00 & 80.5 ± 0.01 & 85.0 ± 0.01 & 84.2 ± 0.02 & 87.8 ± 0.01 \\
VGAE & 91.4 ± 0.01 & 92.6 ± 0.01 & 90.8 ± 0.02 & 92.0 ± 0.02 & 96.4 ± 0.00 &  96.5 ± 0.00 \\
ARGVA & 92.4 ± 0.004 & 93.2 ± 0.003 & 92.4 ± 0.003 & 93.0 ± 0.003 & \textbf{96.8 ± 0.001} & \textbf{97.1 ± 0.001}\\
DGI & 89.8 ± 0.8 & 89.7 ± 1.0 & 95.5 ± 1.0 & 95.7 ± 1.0 & 91.2 ± 0.6 & 92.2 ± 0.5  \\
GIC & 93.5 ± 0.6 & 93.3 ± 0.7 & 97.0 ± 0.5 & 96.8 ± 0.5 & 93.7 ± 0.3 & 93.5 ± 0.3 \\
\hline
\specialrule{0em}{1pt}{1pt}
MNCSCL-CV & \textbf{94.8 ± 0.4} & \textbf{94.2 ± 0.6} & \textbf{97.7 ± 0.4} & \textbf{97.2 ± 0.5} & 94.8 ± 0.2 & 95.4 ± 0.4\\
\specialrule{0em}{0pt}{-0.2pt}
\bottomrule
\end{tabular}
\label{table4}
\end{table*}

\subsection{Implementation Details}

\textbf{Encoder design.} For transductive datasets, we adopt a one-layer Graph Convolutional Network (GCN) as our encoder, with the following propagation rule:
\begin{equation}
\label{eq15}
\mathcal{F}(\textbf{X},\textbf{A})=\sigma(\hat{\textbf{D}}^{-\frac{1}{2}}\hat{\textbf{A}}\hat{\textbf{D}}^{-\frac{1}{2}}\textbf{X}\textbf{W}),
\end{equation}
where $\hat{\textbf{A}}=\textbf{A}+\textbf{I}_N$ is the adjacency matrix with self-loops and $\hat{\textbf{D}}(i,i)=\sum_j\hat{\textbf{A}}(i,j)$ is its corresponding
degree matrix. $\sigma(\cdot)$ is the PReLU nonlinearity \cite{he2015delving} and $\textbf{W}$ is a learnable parameter matrix with $F'=512$ (specially, $F'=256$ on  Pubmed). As for inductive datasets, we adopt a one-layer GCN with skip connections \cite{zhang2019gresnet} as our encoder, with the following propagation rule:
\begin{equation}
\label{eq16}
\mathcal{F}(\textbf{X},\textbf{A})=\sigma(\hat{\textbf{D}}^{-\frac{1}{2}}\hat{\textbf{A}}\hat{\textbf{D}}^{-\frac{1}{2}}\textbf{X}\textbf{W}+\hat{\textbf{A}}\textbf{W}_{skip}),
\end{equation}
where $\textbf{W}_{skip}$ is a learnable parameter matrix with $F'=512$ for skip connections.

\textbf{Corruption function.} For transductive datasets,we transform adjacency matrix $\textbf{A}$ to a diffusion matrix $\textbf{U}$. Specifically, we compute diffusion using fast approximation and sparsification methods \cite{gasteiger2019diffusion} with heat kernel \cite{kondor2002diffusion}:
\begin{equation}
\label{eq-ex}
\textbf{U}=\exp{(t\textbf{AD}^{-1}-t)},
\end{equation}
where $\textbf{D}$ is a diagonal degree matrix as in Section 3.1 and $t$ is diffusion time \cite{gasteiger2019diffusion}. For Reddit dataset, we implement the corruption function $\mathcal{C}$ by keeping the adjacency matrix $\textbf{A}$ unchanged (i.e. $\tilde{\textbf{A}}=\textbf{A}$) and perturbing the feature matrix $\textbf{X}$ by row-wise shuffling. And for PPI dataset, we simply samples a different graph from the training set due to it's a multiple-graph dataset.

\begin{table}[t]
\caption{The classification accuracy (in \%) of different node-centered subgraphs combinations 
(\textit{Subgraph 1}: Basic subgraph, \textit{Subgraph 2}: Neighboring subgraph, \textit{Subgraph 3}: Intimate subgraph, \textit{Subgraph 4}: Communal subgraph, \textit{Subgraph 5}: Full subgraph)
on Cora, Citeseer and Pubmed datasets with same hyperparameter setting. Note that due to the use of the ``core view'' paradigm, there must be at least the basic subgraph and another subgraph. The best result for each dataset is indicated by bolded.}
\centering
\resizebox{1\columnwidth}{!}{
    \begin{tabular}{cccccccc}
    \toprule
    \specialrule{0em}{-0.01pt}{0pt}
    \multicolumn{5}{c}{\textbf{Subgraphs}} & \multicolumn{3}{c}{\textbf{Dataset}} \\
    \specialrule{0em}{0pt}{-0.3pt}
    \cmidrule(r){1-5} \cmidrule(r){6-8}
    \specialrule{0em}{-1pt}{0pt}
    \textbf{1} & \textbf{2} & \textbf{3} & \textbf{4} & \textbf{5} &  \textbf{Cora} & \textbf{Citeseer} & \textbf{Pubmed}\\
    
    \specialrule{0em}{0pt}{-0.2pt}
    \midrule
    \specialrule{0em}{0.6pt}{0.6pt}
    \midrule
    \specialrule{0em}{0.6pt}{0.6pt}
    
    \checkmark & \checkmark & & & &\textit{82.5 ± 0.7} & \textit{72.2 ± 0.7} & \textit{77.5 ± 0.1}\\
    \checkmark & & \checkmark & & &\textit{82.8 ± 0.3} & \textit{72.5 ± 0.7} & \textit{78.6 ± 0.9}\\
    \checkmark & & & \checkmark& &\textit{82.7 ± 0.6} & \textit{72.1 ± 0.6} & \textit{78.2 ± 0.4}\\
    \checkmark & & & & \checkmark &\textit{82.7 ± 0.5} & \textit{71.8 ± 0.6} & \textit{78.1 ± 0.6}\\
    \specialrule{0em}{0pt}{-0.2pt}
    \cmidrule(r){1-5} \cmidrule(r){6-8}
    \specialrule{0em}{-1pt}{0pt}
    \multicolumn{5}{c}{AVG of 2 Subgraphs} & 82.7 ± 0.6 & 72.2 ± 0.7 & 78.2 ± 0.6\\
    
    \specialrule{0em}{0pt}{-0.2pt}
    \midrule
    \specialrule{0em}{0.6pt}{0.6pt}
    \midrule
    \specialrule{0em}{0.6pt}{0.6pt}
    
    \checkmark & \checkmark & \checkmark & & &\textit{83.1 ± 0.5} & \textit{72.7 ± 0.9} & \textit{78.3 ± 0.8}\\
    \checkmark & \checkmark & & \checkmark & &\textit{82.9 ± 0.9} & \textit{72.3 ± 0.5} & \textit{78.0 ± 0.9}\\
    \checkmark & \checkmark & & & \checkmark &\textit{83.2 ± 0.4} & \textit{72.1 ± 0.4} & \textit{77.6 ± 0.8}\\
    \checkmark & & \checkmark & \checkmark & &\textit{82.9 ± 0.5} & \textit{72.0 ± 0.7} & \textit{78.6 ± 0.4}\\
    \checkmark & & \checkmark & & \checkmark &\textit{83.0 ± 0.5} & \textit{72.9 ± 0.5} & \textit{78.7 ± 0.8}\\
    \checkmark & & & \checkmark& \checkmark &\textit{83.1 ± 0.8} & \textit{72.9 ± 0.6} & \textit{78.5 ± 0.6}\\
    \specialrule{0em}{0pt}{-0.2pt}
    \cmidrule(r){1-5} \cmidrule(r){6-8}
    \specialrule{0em}{-1pt}{0pt}
    \multicolumn{5}{c}{AVG of 3 Subgraphs} & 83.0 ± 0.6 & 72.5 ± 0.7 & 78.3 ± 0.8\\

    \specialrule{0em}{0pt}{-0.2pt}
    \midrule
    \specialrule{0em}{0.6pt}{0.6pt}
    \midrule
    \specialrule{0em}{0.6pt}{0.6pt}
    
    \checkmark & \checkmark & \checkmark & \checkmark& &\textit{83.5 ± 0.4} & \textit{72.6 ± 0.7} & \textit{79.2 ± 0.4}\\
    \checkmark & \checkmark & \checkmark & & \checkmark &\textit{83.5 ± 0.4} & \textit{73.0 ± 0.6} & \textit{78.8 ± 0.9}\\
    \checkmark & \checkmark & & \checkmark& \checkmark &\textit{83.6 ± 0.6} & \textit{72.4 ± 0.6} & \textit{78.5 ± 0.8}\\
    \checkmark & & \checkmark & \checkmark& \checkmark &\textit{83.3 ± 0.7} & \textit{73.1 ± 1.0} & \textit{78.6 ± 0.6}\\
    \specialrule{0em}{0pt}{-0.2pt}
    \cmidrule(r){1-5} \cmidrule(r){6-8}
    \specialrule{0em}{-1pt}{0pt}
    \multicolumn{5}{c}{AVG of 4 Subgraphs} & 83.5 ± 0.5 & 72.7 ± 0.8 & 78.7 ± 0.7\\
    \specialrule{0em}{0pt}{-0.2pt}
    \midrule
    \specialrule{0em}{0.6pt}{0.6pt}
    \midrule
    \specialrule{0em}{0.6pt}{0.6pt}
    \checkmark & \checkmark & \checkmark & \checkmark& \checkmark &\textbf{84.7 ± 0.3} & \textbf{73.8 ± 0.5} & \textbf{81.5 ± 0.4}\\

    \specialrule{0em}{0pt}{-0.2pt}
    \midrule
    \specialrule{0em}{0.6pt}{0.6pt}
    \bottomrule
    \specialrule{0em}{0.6pt}{0.6pt}
\end{tabular}}
\label{table5}
\end{table}

\textbf{Readout function.} We use identical readout function $\mathcal{R}$ for all datasets, which performs a simple average of all node features for a given subgraph with $N'$ nodes:
\begin{equation}
\label{eq17}
\mathcal{R}(\textbf{H})=\sigma\left(\frac{1}{N'}\sum_{i=1}^{N'}\textbf{h}_i\right),
\end{equation}
where $\sigma(\cdot)$ is the logistic sigmoid nonlinearity.

\textbf{Discriminator.} We use a simple bilinear scoring
function as discriminator $\mathcal{D}$:
\begin{equation}
\label{eq18}
\mathcal{D}(\textbf{h}_i,\textbf{h}_j)=\sigma(\textbf{h}_i^T\textbf{W}_d\textbf{h}_j), 
\end{equation}
where $\textbf{W}_d$ is a learnable scoring matrix and $\sigma(\cdot)$ is the logistic sigmoid nonlinearity.

\textbf{Details of node-centered subgraphs.} We conduct experiments with all five node-centered subgraphs on two different contrastive losses (named MNCSCL-FG under full graph case and MNCSCL-CV under core view case, respectively). More specifically, we choose $d=1$ for the neighboring subgraph and \textit{Strategy 2} for the communal subgraph. For hyperparameters, we set the size of intimate subgraph $l$ as 20 (specially, 10 on citeseer). The number of clusters $C$ and inverse-temperature hyperparameter $\beta$ for communal subgraph is set to 128 and 10 respectively. To build a proper full subgraph, we also set the self-weighted factor as 0.01.

\subsection{Experimental Results and Analysis}

\textbf{Node classification.} Experiments show that MNCSCL achieves the best performance on all five datasets compared to other competing self-supervised methods, as shown in Table \ref{table3}. We believe this robust performance is due to our comparison of multiple node-centered subgraphs, resulting in learning a more comprehensive node representation. Although MNCSCL-FG and MNCSCL-CV both have shown excellent performance, MNCSCL-FG performs better on PPI dataset and MNCSCL-CV performs better on other datasets. We think this is due to the very sparse available features on the PPI (over 40\% of the nodes have all-zero features). It is more effective to use more perspectives for comparison on such sparse graph datasets. For other datasets, the contrastive loss under core view case is enough for MNCSCL to learn comprehensive information about the nodes, too much comparison will instead lead to overfitting and waste of resources. Compared to supervised methods, MNCSCL also outperforms the two compared methods on all datasets. This shows that our method is also very competitive compared to traditional supervision methods.

\textbf{Link prediction.} To test the generalization capability of MNCSCL, we intend to further investigate the performance of MNCSCL in link prediction task, as shown in Table \ref{table4}. We find that MNCSCL outperforms all competing methods on both the Cora and Citeseer datasets, suggesting that a multiple node-centered subgraphs based comparison can help the model learn node representations with good generalizability. The excellent performance on different downstream tasks further proves the feasibility of our method.

\begin{figure}[t]
\centering
\includegraphics[width=1\columnwidth]{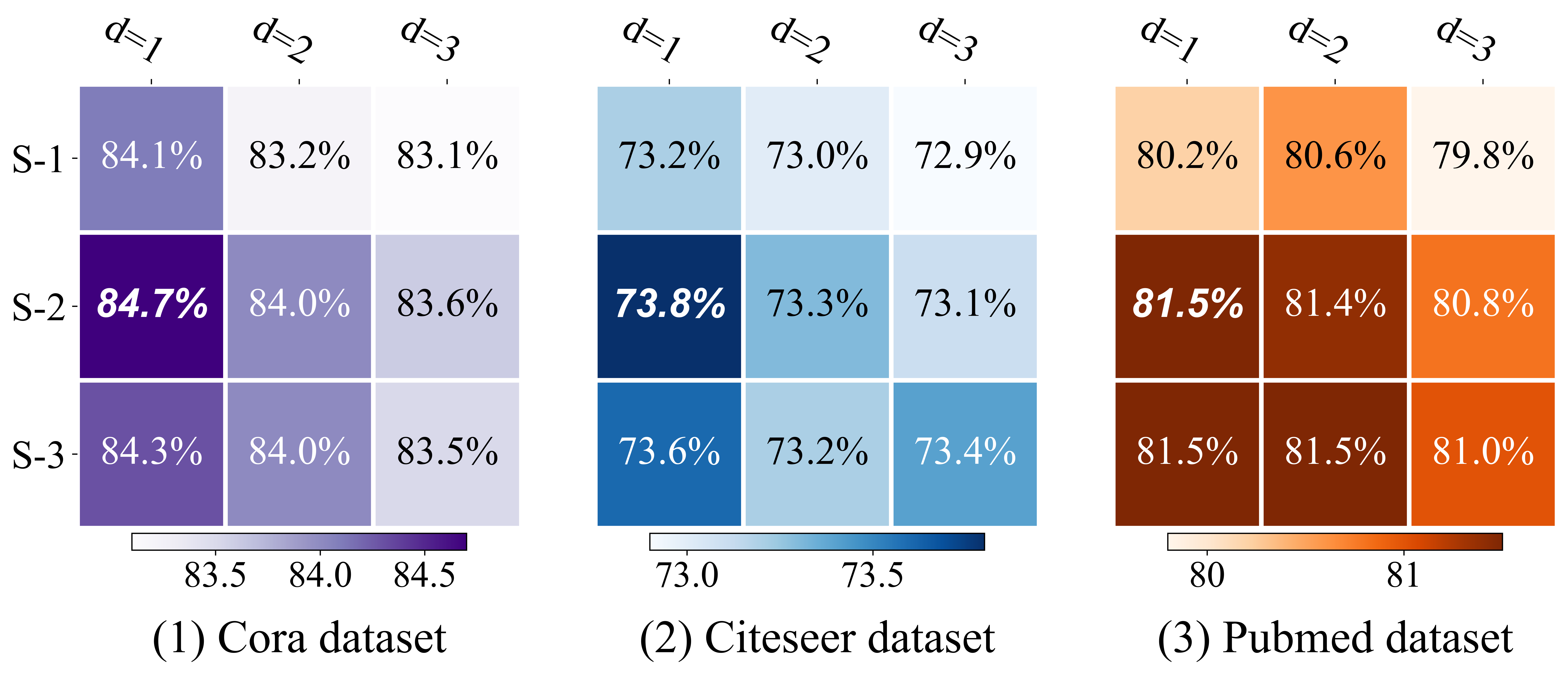} 
\caption{Heat map of classification accuracy (in \%) on three transductive datasets when choosing different range of neighbors in the neighboring subgraph ($d=1$, $d=2$ and $d=3$) and different clustering strategies in the communal subgraph (S-1, S-2 and S-3, where S-1 denotes \textit{Strategy 1}, S-2 denotes \textit{Strategy 2} and S-3 denotes \textit{Strategy 3}). The bolded value in each sub-plot represents the maximum classification accuracy for the current dataset.}
\label{fig5}
\end{figure}

\subsection{Ablation Study}

\textbf{Node-centered subgraph combination.} To investigate how the number of node-centered subgraphs affects the performance of MNCSCL, we permuted and combined five previously mentioned node-centered subgraphs under the core view case and observed the classification accuracy of different subgraph combinations on Cora, Citeseer and Pubmed datasets with same hyperparameter setting, as shown in Table \ref{table5}. Obviously, as the number of node-centered subgraphs increases, the classification accuracy continues to increase, and the best results are achieved when all five subgraphs are used. This indicates that multiple node-centered subgraphs contrastive learning can indeed learn better node representation. 
We also notice that the classification accuracy improvement is more obvious with the increase in the number of node-centered subgraphs.

\textbf{Range of neighbors and clustering strategy.} We investigate the value of $d$ in the neighboring subgraph and the selection of different clustering strategies in the communal subgraph, and the results are shown in Fig \ref{fig5}. Here we use all five node-centered subgraphs with the contrastive loss under core view case. It can be seen that the optimal choice in all three datasets is $d=1$ neighbors and \textit{Strategy 2}. Our analyses are as follows.
\begin{itemize}
    \item For the neighboring subgraph, the classification accuracy tends to decrease as the vale of $d$ increases. We believe that too many neighboring nodes will lead to cause overfitting of the model and performance degradation. This viewpoint can also be verified from the pubmed dataset, where it is not obvious whether the classification accuracy is better or worse when setting $d=1$ and $d=2$. Because the attributes of pubmed are relatively sparse, sometimes its neighboring subgraph needs to contain more neighbors to get better results.
    \item In the choice of clustering strategy, \textit{Strategy 1} is significantly less effective than the other two end-to-end strategies. \textit{Strategy 2} and \textit{Strategy 3} show comparable performance, but considering that \textit{Strategy 3} involves an estimation network, which means more resource consumption, we finally choose \textit{Strategy 2} as the clustering strategy for the communal subgraph.
\end{itemize}

\section{Conclusion}
\label{sec:conclusion}
We propose Multiple Node-centered Subgraphs Contrastive Representation Learning (MNCSCL), a novel approach to self-supervised graph representation learning. MNCSCL obtains five different node-centered subgraphs carefully designed by us through a subgraph generator on each node, and maximizes the mutual information between them through two types of contrastive loss, thus allowing us to obtain comprehensive node representation that combines information from multiple node-centered subgraphs of nodes. Experiments show that MNCSCL reach the advanced level of self-supervised learning in both transductive and inductive node classification tasks as well as in link prediction task.

\begin{acks}
This work is supported by the NSFC program (No. 62272338) and the Natural Science Foundation of Inner Mongolia Autonomous Region of China (No.2022LHMS06008).
\end{acks}

\bibliographystyle{ACM-Reference-Format}
\bibliography{cikm23}


\begin{thebibliography}{42}


\ifx \showCODEN    \undefined \def \showCODEN     #1{\unskip}     \fi
\ifx \showDOI      \undefined \def \showDOI       #1{#1}\fi
\ifx \showISBNx    \undefined \def \showISBNx     #1{\unskip}     \fi
\ifx \showISBNxiii \undefined \def \showISBNxiii  #1{\unskip}     \fi
\ifx \showISSN     \undefined \def \showISSN      #1{\unskip}     \fi
\ifx \showLCCN     \undefined \def \showLCCN      #1{\unskip}     \fi
\ifx \shownote     \undefined \def \shownote      #1{#1}          \fi
\ifx \showarticletitle \undefined \def \showarticletitle #1{#1}   \fi
\ifx \showURL      \undefined \def \showURL       {\relax}        \fi
\providecommand\bibfield[2]{#2}
\providecommand\bibinfo[2]{#2}
\providecommand\natexlab[1]{#1}
\providecommand\showeprint[2][]{arXiv:#2}

\bibitem[Chen et~al\mbox{.}(2018)]%
        {FastGCN}
\bibfield{author}{\bibinfo{person}{Jie Chen}, \bibinfo{person}{Tengfei Ma},
  {and} \bibinfo{person}{Cao Xiao}.} \bibinfo{year}{2018}\natexlab{}.
\newblock \showarticletitle{Fastgcn: fast learning with graph convolutional
  networks via importance sampling}.
\newblock \bibinfo{journal}{\emph{arXiv preprint arXiv:1801.10247}}
  (\bibinfo{year}{2018}).
\newblock


\bibitem[Gasteiger et~al\mbox{.}(2019)]%
        {gasteiger2019diffusion}
\bibfield{author}{\bibinfo{person}{Johannes Gasteiger}, \bibinfo{person}{Stefan
  Wei{\ss}enberger}, {and} \bibinfo{person}{Stephan G{\"u}nnemann}.}
  \bibinfo{year}{2019}\natexlab{}.
\newblock \showarticletitle{Diffusion improves graph learning}.
\newblock \bibinfo{journal}{\emph{Advances in neural information processing
  systems}}  \bibinfo{volume}{32} (\bibinfo{year}{2019}).
\newblock


\bibitem[Grover and Leskovec(2016)]%
        {grover2016node2vec}
\bibfield{author}{\bibinfo{person}{Aditya Grover} {and} \bibinfo{person}{Jure
  Leskovec}.} \bibinfo{year}{2016}\natexlab{}.
\newblock \showarticletitle{node2vec: Scalable feature learning for networks}.
  In \bibinfo{booktitle}{\emph{Proceedings of the 22nd ACM SIGKDD international
  conference on Knowledge discovery and data mining}}.
  \bibinfo{pages}{855--864}.
\newblock


\bibitem[Hamilton et~al\mbox{.}(2017a)]%
        {GraphSAGE}
\bibfield{author}{\bibinfo{person}{Will Hamilton}, \bibinfo{person}{Zhitao
  Ying}, {and} \bibinfo{person}{Jure Leskovec}.}
  \bibinfo{year}{2017}\natexlab{a}.
\newblock \showarticletitle{Inductive representation learning on large graphs}.
\newblock \bibinfo{journal}{\emph{Advances in neural information processing
  systems}}  \bibinfo{volume}{30} (\bibinfo{year}{2017}).
\newblock


\bibitem[Hamilton et~al\mbox{.}(2017b)]%
        {hamilton2017representation}
\bibfield{author}{\bibinfo{person}{William~L Hamilton}, \bibinfo{person}{Rex
  Ying}, {and} \bibinfo{person}{Jure Leskovec}.}
  \bibinfo{year}{2017}\natexlab{b}.
\newblock \showarticletitle{Representation learning on graphs: Methods and
  applications}.
\newblock \bibinfo{journal}{\emph{arXiv preprint arXiv:1709.05584}}
  (\bibinfo{year}{2017}).
\newblock


\bibitem[Hassani and Khasahmadi(2020)]%
        {MVGRL}
\bibfield{author}{\bibinfo{person}{Kaveh Hassani} {and}
  \bibinfo{person}{Amir~Hosein Khasahmadi}.} \bibinfo{year}{2020}\natexlab{}.
\newblock \showarticletitle{Contrastive multi-view representation learning on
  graphs}. In \bibinfo{booktitle}{\emph{International Conference on Machine
  Learning}}. PMLR, \bibinfo{pages}{4116--4126}.
\newblock


\bibitem[He et~al\mbox{.}(2015)]%
        {he2015delving}
\bibfield{author}{\bibinfo{person}{Kaiming He}, \bibinfo{person}{Xiangyu
  Zhang}, \bibinfo{person}{Shaoqing Ren}, {and} \bibinfo{person}{Jian Sun}.}
  \bibinfo{year}{2015}\natexlab{}.
\newblock \showarticletitle{Delving deep into rectifiers: Surpassing
  human-level performance on imagenet classification}. In
  \bibinfo{booktitle}{\emph{Proceedings of the IEEE international conference on
  computer vision}}. \bibinfo{pages}{1026--1034}.
\newblock


\bibitem[Hjelm et~al\mbox{.}(2018)]%
        {DIM}
\bibfield{author}{\bibinfo{person}{R~Devon Hjelm}, \bibinfo{person}{Alex
  Fedorov}, \bibinfo{person}{Samuel Lavoie-Marchildon}, \bibinfo{person}{Karan
  Grewal}, \bibinfo{person}{Phil Bachman}, \bibinfo{person}{Adam Trischler},
  {and} \bibinfo{person}{Yoshua Bengio}.} \bibinfo{year}{2018}\natexlab{}.
\newblock \showarticletitle{Learning deep representations by mutual information
  estimation and maximization}.
\newblock \bibinfo{journal}{\emph{arXiv preprint arXiv:1808.06670}}
  (\bibinfo{year}{2018}).
\newblock


\bibitem[Jeh and Widom(2003)]%
        {jeh2003scaling}
\bibfield{author}{\bibinfo{person}{Glen Jeh} {and} \bibinfo{person}{Jennifer
  Widom}.} \bibinfo{year}{2003}\natexlab{}.
\newblock \showarticletitle{Scaling personalized web search}. In
  \bibinfo{booktitle}{\emph{Proceedings of the 12th international conference on
  World Wide Web}}. \bibinfo{pages}{271--279}.
\newblock


\bibitem[Jiao et~al\mbox{.}(2020)]%
        {SUBG-CON}
\bibfield{author}{\bibinfo{person}{Yizhu Jiao}, \bibinfo{person}{Yun Xiong},
  \bibinfo{person}{Jiawei Zhang}, \bibinfo{person}{Yao Zhang},
  \bibinfo{person}{Tianqi Zhang}, {and} \bibinfo{person}{Yangyong Zhu}.}
  \bibinfo{year}{2020}\natexlab{}.
\newblock \showarticletitle{Sub-graph contrast for scalable self-supervised
  graph representation learning}. In \bibinfo{booktitle}{\emph{2020 IEEE
  international conference on data mining (ICDM)}}. IEEE,
  \bibinfo{pages}{222--231}.
\newblock


\bibitem[Jing and Tian(2020)]%
        {jing2020self}
\bibfield{author}{\bibinfo{person}{Longlong Jing} {and} \bibinfo{person}{Yingli
  Tian}.} \bibinfo{year}{2020}\natexlab{}.
\newblock \showarticletitle{Self-supervised visual feature learning with deep
  neural networks: A survey}.
\newblock \bibinfo{journal}{\emph{IEEE transactions on pattern analysis and
  machine intelligence}} \bibinfo{volume}{43}, \bibinfo{number}{11}
  (\bibinfo{year}{2020}), \bibinfo{pages}{4037--4058}.
\newblock


\bibitem[Ketkar and Moolayil(2021)]%
        {ketkar2021introduction}
\bibfield{author}{\bibinfo{person}{Nikhil Ketkar} {and} \bibinfo{person}{Jojo
  Moolayil}.} \bibinfo{year}{2021}\natexlab{}.
\newblock \showarticletitle{Introduction to pytorch}.
\newblock In \bibinfo{booktitle}{\emph{Deep learning with python}}.
  \bibinfo{publisher}{Springer}, \bibinfo{pages}{27--91}.
\newblock


\bibitem[Kingma and Ba(2014)]%
        {kingma2014adam}
\bibfield{author}{\bibinfo{person}{Diederik~P Kingma} {and}
  \bibinfo{person}{Jimmy Ba}.} \bibinfo{year}{2014}\natexlab{}.
\newblock \showarticletitle{Adam: A method for stochastic optimization}.
\newblock \bibinfo{journal}{\emph{arXiv preprint arXiv:1412.6980}}
  (\bibinfo{year}{2014}).
\newblock


\bibitem[Kipf and Welling(2016a)]%
        {GCN}
\bibfield{author}{\bibinfo{person}{Thomas~N Kipf} {and} \bibinfo{person}{Max
  Welling}.} \bibinfo{year}{2016}\natexlab{a}.
\newblock \showarticletitle{Semi-supervised classification with graph
  convolutional networks}.
\newblock \bibinfo{journal}{\emph{arXiv preprint arXiv:1609.02907}}
  (\bibinfo{year}{2016}).
\newblock


\bibitem[Kipf and Welling(2016b)]%
        {kipf2016variational}
\bibfield{author}{\bibinfo{person}{Thomas~N Kipf} {and} \bibinfo{person}{Max
  Welling}.} \bibinfo{year}{2016}\natexlab{b}.
\newblock \showarticletitle{Variational graph auto-encoders}.
\newblock \bibinfo{journal}{\emph{arXiv preprint arXiv:1611.07308}}
  (\bibinfo{year}{2016}).
\newblock


\bibitem[Kondor and Lafferty(2002)]%
        {kondor2002diffusion}
\bibfield{author}{\bibinfo{person}{Risi~Imre Kondor} {and}
  \bibinfo{person}{John Lafferty}.} \bibinfo{year}{2002}\natexlab{}.
\newblock \showarticletitle{Diffusion kernels on graphs and other discrete
  structures}. In \bibinfo{booktitle}{\emph{Proceedings of the 19th
  international conference on machine learning}}, Vol.~\bibinfo{volume}{2002}.
  \bibinfo{pages}{315--322}.
\newblock


\bibitem[Li et~al\mbox{.}(2018)]%
        {li2018survey}
\bibfield{author}{\bibinfo{person}{Yingming Li}, \bibinfo{person}{Ming Yang},
  {and} \bibinfo{person}{Zhongfei Zhang}.} \bibinfo{year}{2018}\natexlab{}.
\newblock \showarticletitle{A survey of multi-view representation learning}.
\newblock \bibinfo{journal}{\emph{IEEE transactions on knowledge and data
  engineering}} \bibinfo{volume}{31}, \bibinfo{number}{10}
  (\bibinfo{year}{2018}), \bibinfo{pages}{1863--1883}.
\newblock


\bibitem[MacQueen(1967)]%
        {macqueen1967classification}
\bibfield{author}{\bibinfo{person}{J MacQueen}.}
  \bibinfo{year}{1967}\natexlab{}.
\newblock \showarticletitle{Classification and analysis of multivariate
  observations}. In \bibinfo{booktitle}{\emph{5th Berkeley Symp. Math. Statist.
  Probability}}. \bibinfo{pages}{281--297}.
\newblock


\bibitem[Mavromatis and Karypis(2020)]%
        {GIC}
\bibfield{author}{\bibinfo{person}{Costas Mavromatis} {and}
  \bibinfo{person}{George Karypis}.} \bibinfo{year}{2020}\natexlab{}.
\newblock \showarticletitle{Graph infoclust: Leveraging cluster-level node
  information for unsupervised graph representation learning}.
\newblock \bibinfo{journal}{\emph{arXiv preprint arXiv:2009.06946}}
  (\bibinfo{year}{2020}).
\newblock


\bibitem[Mo et~al\mbox{.}(2022)]%
        {SUGRL}
\bibfield{author}{\bibinfo{person}{Yujie Mo}, \bibinfo{person}{Liang Peng},
  \bibinfo{person}{Jie Xu}, \bibinfo{person}{Xiaoshuang Shi}, {and}
  \bibinfo{person}{Xiaofeng Zhu}.} \bibinfo{year}{2022}\natexlab{}.
\newblock \showarticletitle{Simple unsupervised graph representation learning}.
  AAAI.
\newblock


\bibitem[Pan et~al\mbox{.}(2019)]%
        {pan2019learning}
\bibfield{author}{\bibinfo{person}{Shirui Pan}, \bibinfo{person}{Ruiqi Hu},
  \bibinfo{person}{Sai-fu Fung}, \bibinfo{person}{Guodong Long},
  \bibinfo{person}{Jing Jiang}, {and} \bibinfo{person}{Chengqi Zhang}.}
  \bibinfo{year}{2019}\natexlab{}.
\newblock \showarticletitle{Learning graph embedding with adversarial training
  methods}.
\newblock \bibinfo{journal}{\emph{IEEE transactions on cybernetics}}
  \bibinfo{volume}{50}, \bibinfo{number}{6} (\bibinfo{year}{2019}),
  \bibinfo{pages}{2475--2487}.
\newblock


\bibitem[Peng et~al\mbox{.}(2022)]%
        {peng2022towards}
\bibfield{author}{\bibinfo{person}{Qiyao Peng}, \bibinfo{person}{Hongtao Liu},
  \bibinfo{person}{Yinghui Wang}, \bibinfo{person}{Hongyan Xu},
  \bibinfo{person}{Pengfei Jiao}, \bibinfo{person}{Minglai Shao}, {and}
  \bibinfo{person}{Wenjun Wang}.} \bibinfo{year}{2022}\natexlab{}.
\newblock \showarticletitle{Towards a multi-view attentive matching for
  personalized expert finding}. In \bibinfo{booktitle}{\emph{Proceedings of the
  ACM Web Conference 2022}}. \bibinfo{pages}{2131--2140}.
\newblock


\bibitem[Peng et~al\mbox{.}(2020)]%
        {GMI}
\bibfield{author}{\bibinfo{person}{Zhen Peng}, \bibinfo{person}{Wenbing Huang},
  \bibinfo{person}{Minnan Luo}, \bibinfo{person}{Qinghua Zheng},
  \bibinfo{person}{Yu Rong}, \bibinfo{person}{Tingyang Xu}, {and}
  \bibinfo{person}{Junzhou Huang}.} \bibinfo{year}{2020}\natexlab{}.
\newblock \showarticletitle{Graph representation learning via graphical mutual
  information maximization}. In \bibinfo{booktitle}{\emph{Proceedings of The
  Web Conference 2020}}. \bibinfo{pages}{259--270}.
\newblock


\bibitem[Perozzi et~al\mbox{.}(2014)]%
        {deepwalk}
\bibfield{author}{\bibinfo{person}{Bryan Perozzi}, \bibinfo{person}{Rami
  Al-Rfou}, {and} \bibinfo{person}{Steven Skiena}.}
  \bibinfo{year}{2014}\natexlab{}.
\newblock \showarticletitle{Deepwalk: Online learning of social
  representations}. In \bibinfo{booktitle}{\emph{Proceedings of the 20th ACM
  SIGKDD international conference on Knowledge discovery and data mining}}.
  \bibinfo{pages}{701--710}.
\newblock


\bibitem[Qiu et~al\mbox{.}(2020)]%
        {qiu2020gcc}
\bibfield{author}{\bibinfo{person}{Jiezhong Qiu}, \bibinfo{person}{Qibin Chen},
  \bibinfo{person}{Yuxiao Dong}, \bibinfo{person}{Jing Zhang},
  \bibinfo{person}{Hongxia Yang}, \bibinfo{person}{Ming Ding},
  \bibinfo{person}{Kuansan Wang}, {and} \bibinfo{person}{Jie Tang}.}
  \bibinfo{year}{2020}\natexlab{}.
\newblock \showarticletitle{Gcc: Graph contrastive coding for graph neural
  network pre-training}. In \bibinfo{booktitle}{\emph{Proceedings of the 26th
  ACM SIGKDD International Conference on Knowledge Discovery \& Data Mining}}.
  \bibinfo{pages}{1150--1160}.
\newblock


\bibitem[Schaeffer(2007)]%
        {schaeffer2007graph}
\bibfield{author}{\bibinfo{person}{Satu~Elisa Schaeffer}.}
  \bibinfo{year}{2007}\natexlab{}.
\newblock \showarticletitle{Graph clustering}.
\newblock \bibinfo{journal}{\emph{Computer science review}}
  \bibinfo{volume}{1}, \bibinfo{number}{1} (\bibinfo{year}{2007}),
  \bibinfo{pages}{27--64}.
\newblock


\bibitem[Tang and Liu(2011)]%
        {tang2011leveraging}
\bibfield{author}{\bibinfo{person}{Lei Tang} {and} \bibinfo{person}{Huan Liu}.}
  \bibinfo{year}{2011}\natexlab{}.
\newblock \showarticletitle{Leveraging social media networks for
  classification}.
\newblock \bibinfo{journal}{\emph{Data Mining and Knowledge Discovery}}
  \bibinfo{volume}{23}, \bibinfo{number}{3} (\bibinfo{year}{2011}),
  \bibinfo{pages}{447--478}.
\newblock


\bibitem[Tian et~al\mbox{.}(2020)]%
        {CMC}
\bibfield{author}{\bibinfo{person}{Yonglong Tian}, \bibinfo{person}{Dilip
  Krishnan}, {and} \bibinfo{person}{Phillip Isola}.}
  \bibinfo{year}{2020}\natexlab{}.
\newblock \showarticletitle{Contrastive multiview coding}. In
  \bibinfo{booktitle}{\emph{European conference on computer vision}}. Springer,
  \bibinfo{pages}{776--794}.
\newblock


\bibitem[Veli{\v{c}}kovi{\'c} et~al\mbox{.}(2017)]%
        {GAT}
\bibfield{author}{\bibinfo{person}{Petar Veli{\v{c}}kovi{\'c}},
  \bibinfo{person}{Guillem Cucurull}, \bibinfo{person}{Arantxa Casanova},
  \bibinfo{person}{Adriana Romero}, \bibinfo{person}{Pietro Lio}, {and}
  \bibinfo{person}{Yoshua Bengio}.} \bibinfo{year}{2017}\natexlab{}.
\newblock \showarticletitle{Graph attention networks}.
\newblock \bibinfo{journal}{\emph{arXiv preprint arXiv:1710.10903}}
  (\bibinfo{year}{2017}).
\newblock


\bibitem[Velickovic et~al\mbox{.}(2019)]%
        {DGI}
\bibfield{author}{\bibinfo{person}{Petar Velickovic}, \bibinfo{person}{William
  Fedus}, \bibinfo{person}{William~L Hamilton}, \bibinfo{person}{Pietro
  Li{\`o}}, \bibinfo{person}{Yoshua Bengio}, {and} \bibinfo{person}{R~Devon
  Hjelm}.} \bibinfo{year}{2019}\natexlab{}.
\newblock \showarticletitle{Deep Graph Infomax.}
\newblock \bibinfo{journal}{\emph{ICLR (Poster)}} \bibinfo{volume}{2},
  \bibinfo{number}{3} (\bibinfo{year}{2019}), \bibinfo{pages}{4}.
\newblock


\bibitem[Wilder et~al\mbox{.}(2019)]%
        {wilder2019end}
\bibfield{author}{\bibinfo{person}{Bryan Wilder}, \bibinfo{person}{Eric Ewing},
  \bibinfo{person}{Bistra Dilkina}, {and} \bibinfo{person}{Milind Tambe}.}
  \bibinfo{year}{2019}\natexlab{}.
\newblock \showarticletitle{End to end learning and optimization on graphs}.
\newblock \bibinfo{journal}{\emph{Advances in Neural Information Processing
  Systems}}  \bibinfo{volume}{32} (\bibinfo{year}{2019}).
\newblock


\bibitem[Wu et~al\mbox{.}(2019)]%
        {wu2019simplifying}
\bibfield{author}{\bibinfo{person}{Felix Wu}, \bibinfo{person}{Amauri Souza},
  \bibinfo{person}{Tianyi Zhang}, \bibinfo{person}{Christopher Fifty},
  \bibinfo{person}{Tao Yu}, {and} \bibinfo{person}{Kilian Weinberger}.}
  \bibinfo{year}{2019}\natexlab{}.
\newblock \showarticletitle{Simplifying graph convolutional networks}. In
  \bibinfo{booktitle}{\emph{International conference on machine learning}}.
  PMLR, \bibinfo{pages}{6861--6871}.
\newblock


\bibitem[Wu et~al\mbox{.}(2021)]%
        {wu2021self}
\bibfield{author}{\bibinfo{person}{Lirong Wu}, \bibinfo{person}{Haitao Lin},
  \bibinfo{person}{Cheng Tan}, \bibinfo{person}{Zhangyang Gao}, {and}
  \bibinfo{person}{Stan~Z Li}.} \bibinfo{year}{2021}\natexlab{}.
\newblock \showarticletitle{Self-supervised learning on graphs: Contrastive,
  generative, or predictive}.
\newblock \bibinfo{journal}{\emph{IEEE Transactions on Knowledge and Data
  Engineering}} (\bibinfo{year}{2021}).
\newblock


\bibitem[Xu et~al\mbox{.}(2021)]%
        {xu2021self}
\bibfield{author}{\bibinfo{person}{Minghao Xu}, \bibinfo{person}{Hang Wang},
  \bibinfo{person}{Bingbing Ni}, \bibinfo{person}{Hongyu Guo}, {and}
  \bibinfo{person}{Jian Tang}.} \bibinfo{year}{2021}\natexlab{}.
\newblock \showarticletitle{Self-supervised graph-level representation learning
  with local and global structure}. In \bibinfo{booktitle}{\emph{International
  Conference on Machine Learning}}. PMLR, \bibinfo{pages}{11548--11558}.
\newblock


\bibitem[Zhang and Meng(2019)]%
        {zhang2019gresnet}
\bibfield{author}{\bibinfo{person}{Jiawei Zhang} {and} \bibinfo{person}{Lin
  Meng}.} \bibinfo{year}{2019}\natexlab{}.
\newblock \showarticletitle{Gresnet: Graph residual network for reviving deep
  gnns from suspended animation}.
\newblock \bibinfo{journal}{\emph{arXiv preprint arXiv:1909.05729}}
  (\bibinfo{year}{2019}).
\newblock


\bibitem[Zhang et~al\mbox{.}(2020)]%
        {zhang2020graph}
\bibfield{author}{\bibinfo{person}{Jiawei Zhang}, \bibinfo{person}{Haopeng
  Zhang}, \bibinfo{person}{Congying Xia}, {and} \bibinfo{person}{Li Sun}.}
  \bibinfo{year}{2020}\natexlab{}.
\newblock \showarticletitle{Graph-bert: Only attention is needed for learning
  graph representations}.
\newblock \bibinfo{journal}{\emph{arXiv preprint arXiv:2001.05140}}
  (\bibinfo{year}{2020}).
\newblock


\bibitem[Zhao(2021)]%
        {zhao2021fairnessphd}
\bibfield{author}{\bibinfo{person}{Chen Zhao}.}
  \bibinfo{year}{2021}\natexlab{}.
\newblock \emph{\bibinfo{title}{Fairness-Aware Multi-Task and Meta Learning}}.
\newblock \bibinfo{thesistype}{Ph.\,D. Dissertation}.
\newblock


\bibitem[Zhao and Chen(2019)]%
        {zhao2019rank}
\bibfield{author}{\bibinfo{person}{Chen Zhao} {and} \bibinfo{person}{Feng
  Chen}.} \bibinfo{year}{2019}\natexlab{}.
\newblock \showarticletitle{Rank-based multi-task learning for fair
  regression}. In \bibinfo{booktitle}{\emph{2019 IEEE International Conference
  on Data Mining (ICDM)}}. IEEE, \bibinfo{pages}{916--925}.
\newblock


\bibitem[Zhao et~al\mbox{.}(2021)]%
        {zhao2021fairness}
\bibfield{author}{\bibinfo{person}{Chen Zhao}, \bibinfo{person}{Feng Chen},
  {and} \bibinfo{person}{Bhavani Thuraisingham}.}
  \bibinfo{year}{2021}\natexlab{}.
\newblock \showarticletitle{Fairness-aware online meta-learning}. In
  \bibinfo{booktitle}{\emph{Proceedings of the 27th ACM SIGKDD Conference on
  Knowledge Discovery \& Data Mining}}. \bibinfo{pages}{2294--2304}.
\newblock


\bibitem[Zhao et~al\mbox{.}(2020)]%
        {zhao2020multi}
\bibfield{author}{\bibinfo{person}{Jun Zhao}, \bibinfo{person}{Xudong Liu},
  \bibinfo{person}{Qiben Yan}, \bibinfo{person}{Bo Li},
  \bibinfo{person}{Minglai Shao}, {and} \bibinfo{person}{Hao Peng}.}
  \bibinfo{year}{2020}\natexlab{}.
\newblock \showarticletitle{Multi-attributed heterogeneous graph convolutional
  network for bot detection}.
\newblock \bibinfo{journal}{\emph{Information Sciences}}  \bibinfo{volume}{537}
  (\bibinfo{year}{2020}), \bibinfo{pages}{380--393}.
\newblock


\bibitem[Zhu et~al\mbox{.}(2020)]%
        {GRACE}
\bibfield{author}{\bibinfo{person}{Yanqiao Zhu}, \bibinfo{person}{Yichen Xu},
  \bibinfo{person}{Feng Yu}, \bibinfo{person}{Qiang Liu}, \bibinfo{person}{Shu
  Wu}, {and} \bibinfo{person}{Liang Wang}.} \bibinfo{year}{2020}\natexlab{}.
\newblock \showarticletitle{Deep graph contrastive representation learning}.
\newblock \bibinfo{journal}{\emph{arXiv preprint arXiv:2006.04131}}
  (\bibinfo{year}{2020}).
\newblock


\bibitem[Zimmermann and Nagappan(2008)]%
        {zimmermann2008predicting}
\bibfield{author}{\bibinfo{person}{Thomas Zimmermann} {and}
  \bibinfo{person}{Nachiappan Nagappan}.} \bibinfo{year}{2008}\natexlab{}.
\newblock \showarticletitle{Predicting defects using network analysis on
  dependency graphs}. In \bibinfo{booktitle}{\emph{Proceedings of the 30th
  international conference on Software engineering}}.
  \bibinfo{pages}{531--540}.
\newblock


\end{thebibliography}

\appendix

\end{document}